\documentclass[runningheads]{llncs}

\usepackage{eccv}

\usepackage{eccvabbrv}

\usepackage{graphicx}
\usepackage{booktabs}
\usepackage{multirow}
\usepackage{wrapfig, lipsum, booktabs}
\usepackage{makecell}

\usepackage{color, colortbl}
\definecolor{lightgray}{gray}{.9}

\usepackage[accsupp]{axessibility}  % Improves PDF readability for those with disabilities.
\newcommand{\method}{AnyControl\xspace}
\newcommand{\encname}{Multi-Control Encoder\xspace}
\newcommand{\topone}[1]{\textbf{\textcolor{red}{#1}}}
\newcommand{\toptwo}[1]{\textbf{\textcolor{blue}{#1}}}
\newcommand{\benchmark}{COCO-UM\xspace}

\usepackage[pagebackref,breaklinks,colorlinks,citecolor=eccvblue]{hyperref}
\usepackage{orcidlink}
\usepackage{comment}
\newcommand\correspondingauthor{\thanks{Corresponding author.}}

\begin{document}

% ---------------------------------------------------------------
% TODO REVIEW: Replace with your title
\title{\method: Create Your Artwork with Versatile Control on Text-to-Image Generation} 

% TODO REVIEW: If the paper title is too long for the running head, you can set
% an abbreviated paper title here. If not, comment out.
\titlerunning{Abbreviated paper title}

% TODO FINAL: Replace with your author list. 
% Include the authors' OCRID for the camera-ready version, if at all possible.
\author{Yanan Sun\inst{1} \and
Yanchen Liu\inst{1,2} \and
Yinhao Tang\inst{1} \and
Wenjie Pei\inst{2} \and
Kai Chen\inst{1}\correspondingauthor
}

\authorrunning{Sun. et al.}
% First names are abbreviated in the running head.
% If there are more than two authors, 'et al.' is used.

% TODO FINAL: Replace with your institution list.
\institute{Shanghai AI Laboratory
 \and
Harbin Institute of Technology, Shenzhen
\\
\email{now.syn@gmail.com} \\
\email{wenjiecoder@outlook.com} \\
\email{\{liuyanchen, tangyinhao, chenkai\}@pjlab.org.cn}
}

{
\maketitle
\begin{center}
\vspace{-0.1in}
    \centering
    \captionsetup{type=figure}
    \includegraphics[width=1.0\textwidth]{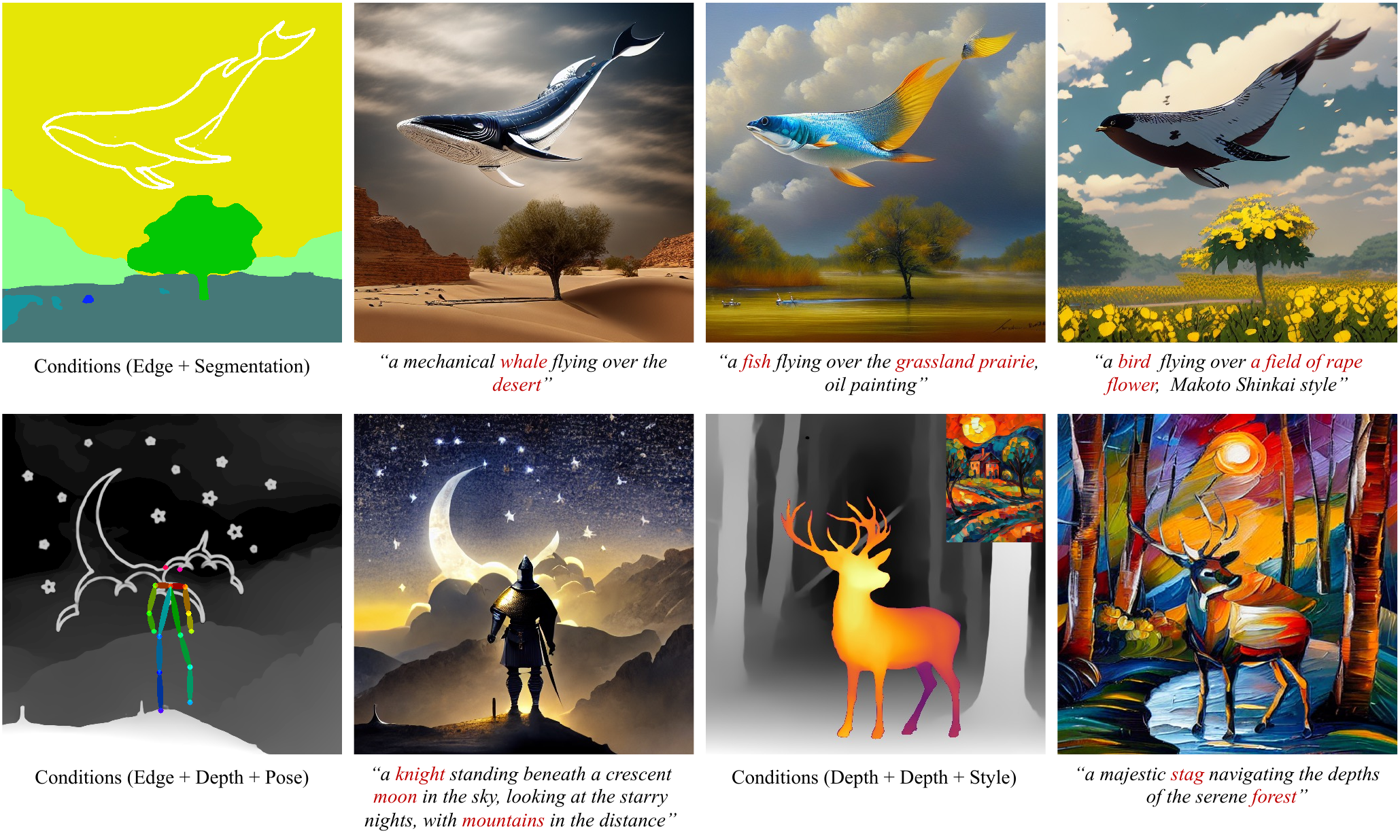}
    \vspace{-18pt}
    \captionof{figure}{\label{fig:teaser}
    Multi-control image synthesis of \textbf{\method}. 
    Our model supports free combinations of multiple control signals and generates harmonious results that are well-aligned with each input. The input control signals fed into the model are shown in a combined image for better visualization.
    }
\end{center}
}

\vspace{-0.2in}
\begin{abstract}
The field of text-to-image (T2I) generation has made significant progress in recent years, largely driven by advancements in diffusion models. Linguistic control enables effective content creation, but struggles with fine-grained control over image generation. This challenge has been explored, to a great extent, by incorporating additional user-supplied spatial conditions, such as depth maps and edge maps, into pre-trained T2I models through extra encoding. However, multi-control image synthesis still faces several challenges.
Specifically, current approaches are limited in handling free combinations of diverse input control signals, overlook the complex relationships among multiple spatial conditions, and often fail to maintain semantic alignment with provided textual prompts. 
This can lead to suboptimal user experiences.
To address these challenges, we propose \textbf{\method}, a multi-control image synthesis framework that supports arbitrary combinations of diverse control signals. \method develops a novel Multi-Control Encoder that extracts a unified multi-modal embedding to guide the generation process. 
This approach enables a holistic understanding of user inputs, and produces high-quality, faithful results under versatile control signals, as demonstrated by extensive quantitative and qualitative evaluations. Our project page is available in \url{https://any-control.github.io}.

\vspace{-0.1in}
\end{abstract}

\vspace{-0.25in}
\section{Introduction}
\label{sec:intro}
\vspace{-0.1in}

In recent years, the field of text-to-image (T2I) generation has experienced significant advancements, leading to unprecedented improvements in generated image quality and diversity, primarily attributed to the introduction of diffusion models~\cite{DDPM, DDIM, CFG, LDM}. While linguistic control allows for effective and engaging content creation, it also presents challenges in achieving fine-grained control over image generation. This challenge is extensively explored in~\cite{ControlNet, T2I-Adapter}, where an additional network is employed to encode and inject the user-supplied control signal into the pre-trained T2I model such as Stable Diffusion~\cite{LDM}, so that to exert influence over the image generation process. Built upon~\cite{ControlNet}, subsequent approaches~\cite{unicontrol, uni-controlnet, Cocktail} presents unified architecture designs for managing multiple spatial conditions. 

However, the task of \textbf{multi-control image synthesis} remains challenging in the following aspects: (1) accommodating free combinations of input conditions, (2) modeling complex relationships among multiple spatial conditions, and (3) maintaining compatibility with textual prompts. We refer to these three challenges as input flexibility, spatial compatibility, and textual compatibility, respectively.

\textbf{Input flexibility}. The first challenge comes from the any combination of available control signals based on user requirements. The amount and modality of control signals provided by users are varying, placing high demands on the input flexibility of the model. However, existing methods~\cite{uni-controlnet, Cocktail} typically employ fixed-length input channels, limiting their ability to accommodate diverse inputs. Other approaches~\cite{ControlNet,T2I-Adapter,unicontrol} adopt MoE design to solve varying-number conditions, which can result in unforeseen artifacts when processing unseen combinations of inputs.

\textbf{Spatial compatibility}. Secondly, control signals are not isolated; instead, they collectively influence the composition of a complete image. It is crucial to consider the relationships among these control signals, especially when managing occlusions among multiple spatial conditions. Unfortunately, current algorithms commonly combine multiple conditions through weighted summation with hand-crafted weights, easily leading to undesired blending results, or even causing low-response control signals to disappear when addressing occlusions. 

\textbf{Textual compatibility}. Ultimately, textual compatibility emerges as an important factor influencing user experience.  Typically, the textual descriptions govern the content of generated images, whereas spatial conditions compensate the structural information.  Nevertheless, a lack of communication between the textual and spatial conditions often leads current algorithms to prioritize accommodating the spatial conditions, thereby disregarding the impact of textual prompts.

In summary, generating comprehensive and harmonious results that satisfy both textual prompts and multiple spatial conditions presents a significant challenge for multi-control image synthesis. To tackle the challenges of input flexibility, spatial compatibility, and textual compatibility, we propose \textbf{\method}, a controllable image synthesis framework that supports arbitrary combinations of diverse control signals.

At the core of \method is the \textbf{Multi-Control Encoder}, which plays a crucial role in ensuring coherent, spatially and semantically aligned multi-modal embeddings. This novel component allows \method to extract a unified representation from various control signals, enabling a truly versatile and high-performing multi-control image synthesis framework. 

Specifically, Multi-Control Encoder is driven by \textbf{multi-control fusion} block and  \textbf{multi-control alignment} block in turns, with a set of \textbf{query tokens} to unite the two seamlessly. 

Multi-control fusion block is employed to aggregate compatible information from multiple spatial conditions through the query tokens. A cross-attention transformer block is employed on the query tokens and the visual tokens of spatial conditions extracted from a pre-trained visual encoder. Therefore, the rich spatial controllable information is passed to the query tokens, which will be utilized in the multi-control alignment block.

Multi-control alignment is used to guarantee the compatibility between all forms of control signals by aligning all other signals to the textual signal. A self-attention transformer block is employed on the query tokens and textual tokens. The query tokens contain spatial controllable information, while the textual tokens carry semantic information. Through information exchange between the query and textual tokens, both types of tokens are able to represent compatible multi-modal information.

With alternating multi-control fusion and alignment blocks in several turns, the query tokens achieve a comprehensive understanding with highly aligned and compatible information from versatile user inputs.  This capability empowers our method to handle complex relationships among conditions and uphold strong compatibility with textual prompts.
Consequently, this approach fosters a more smooth and harmonious control over the generated images. Furthermore, transformer blocks with attention mechanisms inherently excel in accommodating a variety of control signals, and thus enable free combinations of user inputs.

In summary, our contributions are manifold in the following:
\begin{enumerate}
\vspace{-0.05in}
    \item \method proposes a novel Multi-Control Encoder comprising a sequence of alternative multi-control fusion and alignment blocks to achieve comprehensive understanding of complex multi-modal user inputs. 
    \item \method supports flexible combinations of user inputs, regardless of the amount and modality of different control signals.
    \item \method produces more harmonious and natural high-quality outcomes, demonstrating state-of-the-art performance in multi-control image synthesis.
\end{enumerate}

\section{Related Work}

\subsection{Text-to-image Generation}
T2I diffusion models~\cite{LDM,DALLE2,Imagen,GLIDE} have emerged as a promising approach for generating high-quality images from textual prompts. Diffusion models~\cite{DDPM, DDIM}, originally developed for image generation, have been adapted to the T2I domain, offering a novel perspective on the problem. These models leverage the concept of iterative denoising, where the generation process unfolds step-by-step, progressively refining the image quality. The diffusion process allows for better control over the generated images by conditioning on both the text input and intermediate image representations at each diffusion step. Recent advances in T2I diffusion models have explored various techniques to enhance the generation process, such as introducing attention mechanisms to better align textual and visual features~\cite{LDM} and operating in latent space~\cite{LDM, DALLE2} to achieve complexity reduction and detail preservation. 
While T2I diffusion models have shown promising results, there is still ongoing research to address challenges such as controllability, also the focus discussed in this paper, in the context of diffusion-based T2I generation.

\vspace{-0.15in}
\subsection{Controllable Image Synthesis}
Text descriptions guide the diffusion model to generate user-desired images but are insufficient in fine-grained control over the generated results. The fine-grained control signals are diverse in modality, for instance, layout constraint is introduced to arrange the location of the objects given; a bundle of works~\cite{LayoutDiffuse, LayoutDiffusion, SSMG, Law-diffusion, Layoutllm-t2i, BoxDiff, GLIGEN, ReCO} thoroughly explore to synthesis images with high layout alignment given the semantic-aware boxes. Besides, segmentation map~\cite{ZestGuide, make-a-scene, SpaText, MultiDiffusion} is another popular control signal for controlling the layout and object shape of generated images. InstanceFusion~\cite{InstanceDiffusion} proposes a method to support location control in more free-form such as point, scribble, box and segmentation map. Highly detailed control can be achieved by structure signal such as sketch~\cite{MaskSketch}. Depth map~\cite{CnC} can provide the control of the depth of field for the generated images. As layout, structure and depth control all outline the generated image in spatial alignment, content control~\cite{TextualInversion, DreamBooth, IP-Adapter} enables the personalization of the generated appearance in semantic level through an additional image input. 

Studies~\cite{T2I-Adapter, ControlNet, FreeControl} propose general framework designs to process diverse spatial conditions 
% by directly switching the modality of training dataset 
instead of control-specific design, while both spatial and content control are jointly taken into consideration in works~\cite{Composer, DiffBlender}. Specifically, considering the powerful generation ability of T2I model, ControlNet~\cite{ControlNet} proposes to utilize the trainable copy of the UNet encoder in the T2I diffusion model to encode extra condition signals into latent representations and then apply zero convolution to inject into the backbone of the UNet in diffusion modal. The simple but effective design shows generalized and stable performance in spatial control, and thus are widely adopted in various downstream applications. However, ControlNet is a single-modality framework and requires separate model for each modality. To address this, unified ControlNet-like models~\cite{unicontrol, uni-controlnet, Cocktail} are proposed to handle diverse control signals with only one multi-modality model. Another advantage of these methods is that they can support multi-control image synthesis.  They adopt fixed length input channels or MoE design with hand-crafted weighted summation to aggregate conditions. Nevertheless, these methods are short in handling conditions with complex relations and hard to generate harmonious, natural results under various control signals.

\begin{figure}[t]
  \centering
  \vspace{-0.05in}
  \includegraphics[width=1.\linewidth]{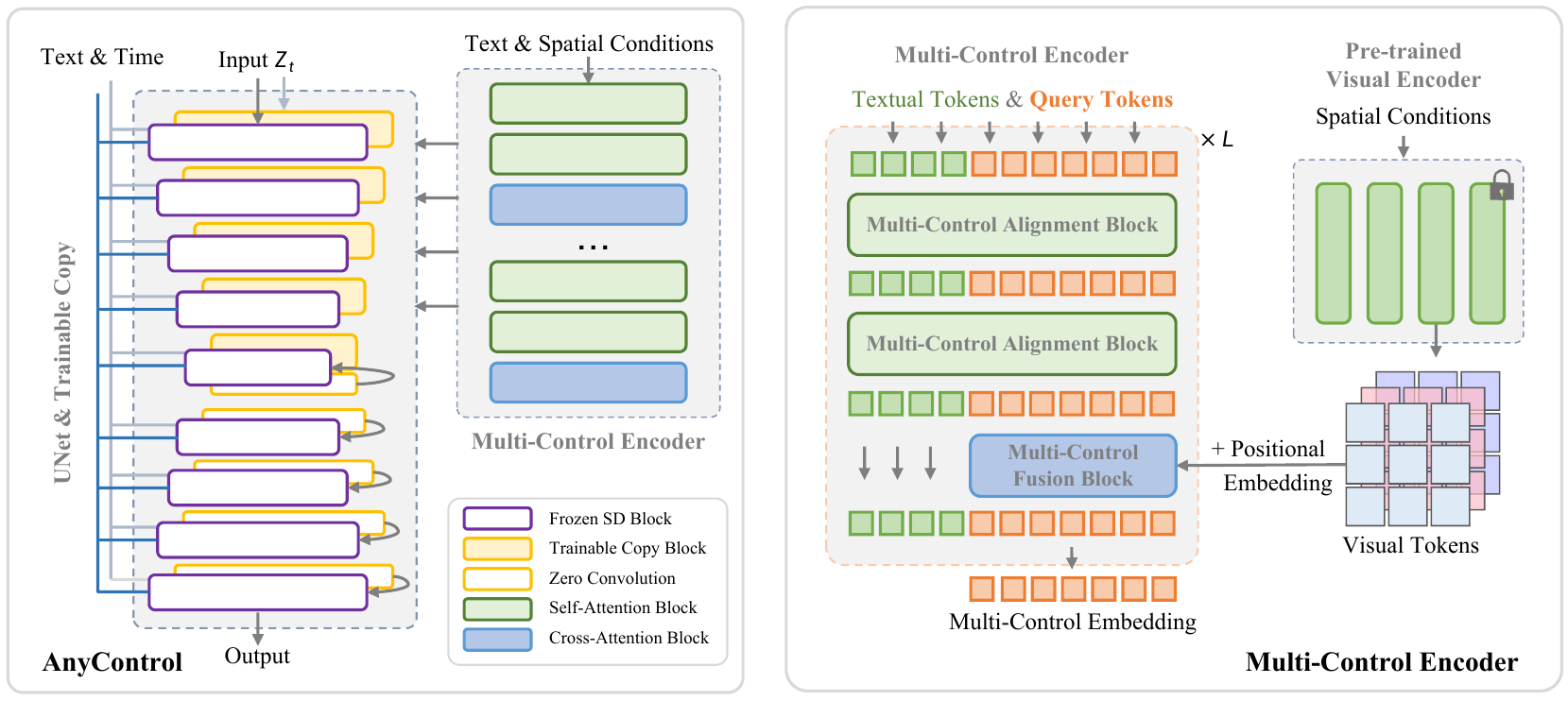}
  \caption{\textbf{AnyControl and Multi-Control Encoder.} \textbf{Left} shows the overall framework of our AnyControl, which develops a Multi-Control Encoder for extracting comprehensive multi-control embeddings based on the textual prompts and multiple spatial conditions. The multi-control embeddings are then utilized to guide the generation process. \textbf{Right} shows the detailed design of our Multi-Control Encoder driven by alternating multi-control fusion and alignment blocks, with query tokens defined to distill the compatible information from textual tokens and visual tokens of the spatial conditions.}
  \label{fig:framework}
  \vspace{-0.25in}
\end{figure}

\vspace{-0.1in}
\section{Method}
In this section, we first give a preliminary overview of Stable Diffusion~\cite{LDM} and ControlNet~\cite{ControlNet}. Subsequently, we introduce \method, featuring a pioneering \encname crafted for extracting a unified representation with compatible information for multiple control signals. Finally, we expound on our training dataset and strategy. Figure~\ref{fig:framework} depicts the architecture of AnyControl and the Multi-Control Encoder.

\vspace{-0.1in}
\subsection{Preliminary}
\textbf{Stable Diffusion.} 
T2I generation introduces text as conditions in diffusion models. In the forward pass, Gaussian noises are gradually added to the sample over a series of steps; while the backward process learns to recover the image by estimating and eliminating the noise with the text guidance.
In this paper, we base Stable Diffusion~\cite{LDM}, one of the most popular T2I diffusion model, to develop \method for multi-control image synthesis.

Stable Diffusion model operates the diffusion and denoising process in latent space rather than pixels to reduce computation cost. It adopts UNet-like~\cite{UNet} structure as its backbone, comprising downsampling blocks, middle block and upsampling blocks.
The text guidance are encoded through CLIP~\cite{CLIP} text encoder and integrated into the UNet through a \textit{CrossAttention} block after each ResBlock~\cite{ResNet}.  If we use $Z$ to denote the noise features derived from the last ResBlock and $Y$ to denote the embeddings encoded by the text encoder, the output noise features $\Tilde{Z}$ from \textit{CrossAttention} block can be obtained by
\begin{gather}
    Q =  W_q(Z), K = W_k(Y), V = W_v(Y), \\
    \Tilde{Z} = Softmax(\frac{QK^T}{\sqrt{d}})V,
\end{gather}
where $W_q$, $W_k$ and $W_v$ are projection layer and $d$ is the dimension of the embedding space.

\vspace{0.1in}
\noindent{\textbf{ControlNet.}}
ControlNet~\cite{ControlNet} is developed to adapt Stable Diffusion model for spatial conditions. To be specific, it locks the parameters of Stable Diffusion, and makes a trainable copy of the encoding layers in the UNet. The two parts are connected by zero convolution layers with zero-initialized weights to progressively increase spatial control influence as the training goes. This design empowers ControlNet to achieve robust controllable image generation while preserve the quality and capabilities of Stable Diffusion model.

\subsection{\method}
\vspace{0.1in}
\noindent{\textbf{\encname.}} 
Similar to ControlNet, in our \method, we also lock the pre-trained stable diffusion model, and instead design a \encname for understanding complex control signals. We first obtain three types of tokens, i.e., textual tokens $\mathcal{T}$, visual tokens $\mathcal{V}$ and query tokens $\mathcal{Q}$. The textual tokens are extracted from CLIP text encoder on textual prompts,  while the visual tokens are obtained from a pre-trained visual encoder (e.g., CLIP image encoder) on all of the user-provided spatial conditions in image form. The query tokens are defined as a set of learnable parameters. 
To address the three challenges discussed in the introduction, \ie, input flexibility, spatial compatibility and textual compatibility, we develop the multi-control encoder via alternating multi-control fusion blocks and multi-control alignment blocks united by the query tokens.

\vspace{0.1in}
\noindent{\textbf{Multi-Control Fusion.}} Multi-control fusion block aims to extract compatible information from various spatial conditions. This is accomplished by utilizing a cross-attention transformer block to facilitate interactions between the query tokens and the visual tokens of all spatial conditions.

Specifically, suppose that there are $n$ spatial conditions in image form of various modalities including depth, segmentation, etc. We can obtain the visual tokens $\mathcal{V}_{i,j}$ for condition $C_i$ from the $j$-th block of the pre-trained visual encoder. Here, we use $[\mathcal{V}_{1,j}, \mathcal{V}_{2,j}, \dots, \mathcal{V}_{n,j}]$ to represent the visual tokens for all the spatial conditions from the $j$-th block. Then the interactions in the multi-control fusion block can be formulated as
\begin{align}
    \mathcal{Q}_j = CrossAttention(\mathcal{Q}_j, [\mathcal{V}_{1,j}+P, \mathcal{V}_{2,j}+P, \dots, \mathcal{V}_{n,j}+P]),
\end{align}
where $P$ denotes a shared learnable positional embedding additive to each $\mathcal{V}_{i,j}$ for better alignment between the query tokens and the visual tokens.

After this process, the spatial controllable information encoded in the visual tokens is passed on to the query tokens.

\vspace{0.1in}
\noindent{\textbf{Multi-Control Alignment.}} Although the various controllable information is integrated into the query tokens, it is challenging to infer the priority of spatial control signals within the overlapping region due to the absence of a global condition that indicates the relationships among spatial conditions. Fortunately, textual prompts can serve as a global control that regulates the content of generated image. Therefore, in the multi-control alignment block, we facilitate the interactions between the query tokens and the textual tokens with a self-attention transformer block. Before we encode the textual prompts to tokens, we append a textual task prompt at the tail of the user-provided text to solve the modality discrepancy among diverse spatial conditions. Then we concatenate the query tokens $\mathcal{Q}$ and textual tokens $\mathcal{T}$ together and perform the self-attention as 
\begin{align}
    [\mathcal{Q}_{j+1}, \mathcal{T}_{j+1}] = SelfAttention([\mathcal{Q}_j, \mathcal{T}_j]).
\end{align}
With self-attention, the query tokens, which carry the mixed controllable information, will exchange information with textual tokens and thus can achieve semantic alignment with user prompts. 

\vspace{0.1in}
\noindent{\textbf{Alternating Fusion and Alignment.}} To ensure the information aligned and compatible of all the control signals, we employ the multi-control fusion and alignment blocks alternately for multiple turns. Notably, we utilize multi-level visual tokens for fine-grained spatial control. Specifically, in each turn, the visual tokens consumed in the cross-attention transformer block are extracted from different levels of the pre-trained visual encoder, considering the spatial conditions are diverse in controlling level, \ie, layout control such as segmentation map and structural control such as edge map. Therefore, multi-level visual tokens are necessary for multi-control fusion blocks in different depth.

\vspace{0.1in}
\noindent{\textbf{Advantages of AnyControl.}} The query tokens work as a bridge, uniting the two types of blocks seamlessly. After several turns, the query tokens retain well-aligned compositional information, served as a unified multi-modal representations for user inputs. This design empowers AnyControl in multi-control image synthesis even with occlusion, generating high-quality harmonious results with high spatial and textual compatibility. Our multi-control encoder shares a similar idea to Q-Former~\cite{BLIP2}, however, AnyControl incorporates many dedicated design for the multi-control image synthesis such as the appended textual task prompt, additional shared position embeddings across all the conditions and the usage of multi-level visual tokens. In implementation, to save computation cost, we insert the cross-attention block after every two self-attention blocks.

\begin{figure}[t]
    \centering
    \includegraphics[width=0.85\linewidth]{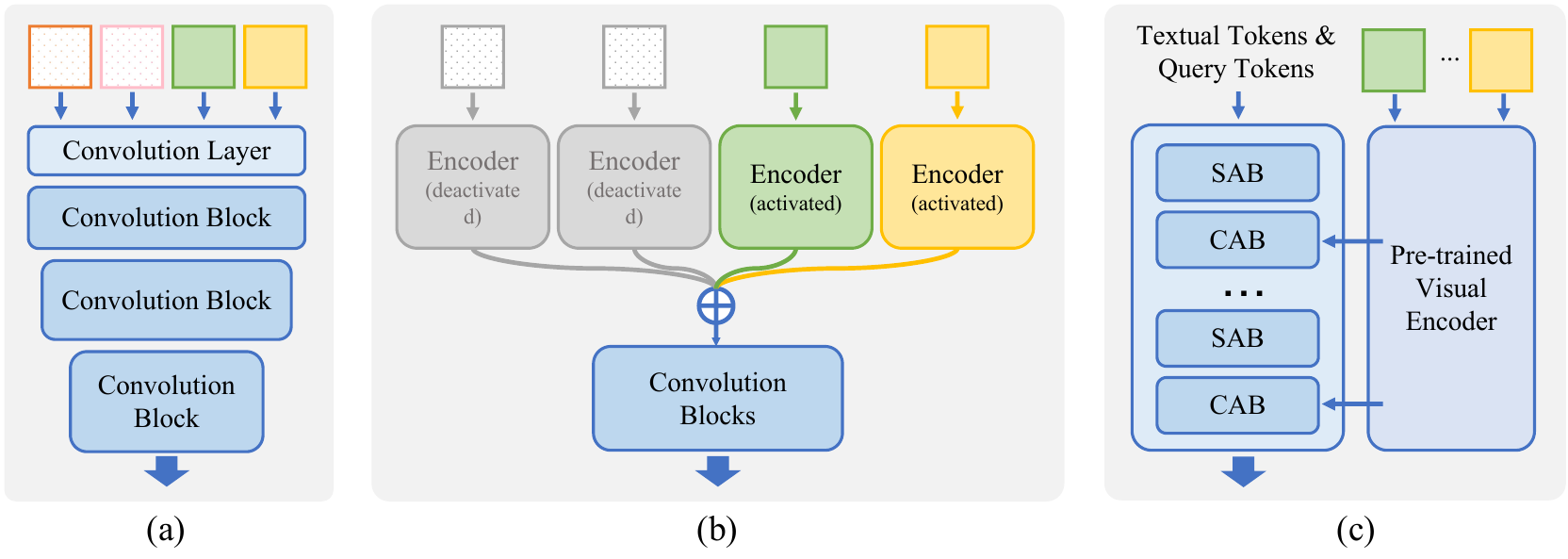}
    \vspace{-0.1in}
    \caption{\textbf{Three types of multi-control methods.} Square in different color denotes different condition type while dotted square denotes zero tensor. \textbf{(a)} Some methods~\cite{uni-controlnet, Cocktail} adopt fixed-length channels of the input convolution layer followed by several convolution blocks to serve as the Multi-Control Encoder. \textbf{(b)} Other methods~\cite{unicontrol, ControlNet, T2I-Adapter} utilize the MoE design, that is, construct separate encoder for each type of control signal and then obtain the embeddings through weighted sum. \textbf{(c)} Different from them, AnyControl adopts attention mechanism to accommodate varying-number and varying-modality of conditions. ``SAB'' and ``CAB'' denotes self- and cross-attention block, respectively. }
    \label{fig:different_methods}
    \vspace{-0.3in}
\end{figure}

Another natural advantage of our AnyControl lies in the input flexibility.  AnyControl, utilizing the transformer blocks with attention mechanism, has natural advantage in accommodating free combinations of user inputs. Previous methods either adopt the design of fixed-length input channels or MoE structure as illustrated in Figure~\ref{fig:different_methods}. The former limits the freedom of user inputs, while the latter, MoE design, supports combining flexible inputs with hand-crafted weighted summation, leading to laborious adjustments to the combination weights.

\vspace{-0.11in}
\subsection{Training}\label{sec:training}
\noindent{\textbf{Datasets.}} 
\begin{wrapfigure}{r}{0.63\textwidth}
\centering
\vspace{-0.5in}
\includegraphics[width=0.62\textwidth]{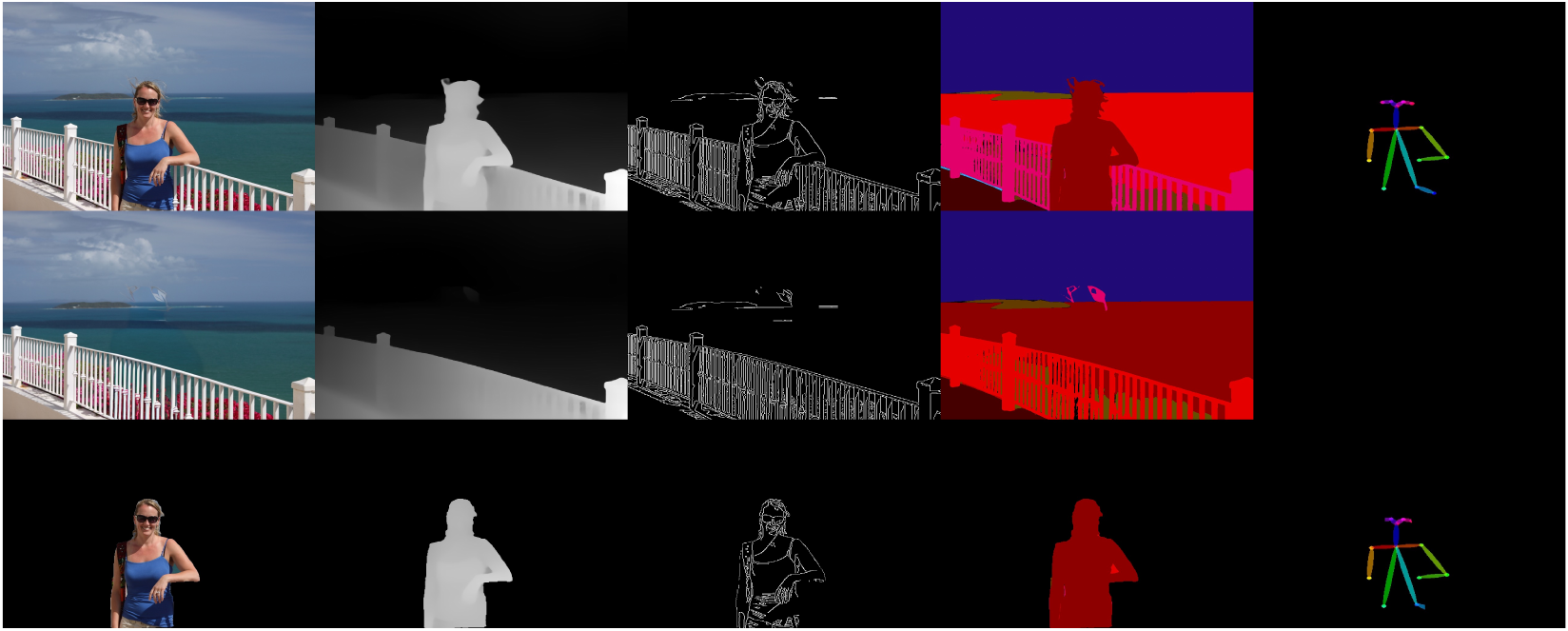}
\vspace{-0.1in}
\caption{\textbf{Visualization of aligned and unaligned conditions.}  The first row shows the aligned case where pixels at the same location of all the control signals describe the same object. Conditions in the second and third rows describe the foreground and background respectively, contributing to a complete image together, constructing the unaligned case.}
\label{fig:unaligned_data}
\vspace{-0.4in}
\end{wrapfigure}
We adopt the 
training dataset, MultiGen, for multi-control image synthesis presented in~\cite{unicontrol}. 
This dataset is built from LAION~\cite{laion} with aesthetics score above 6. Low-resolution images are removed and finally 2.8M images are kept. Different methods are utilized to extract the control signals. 
Unfortunately, there is a domain gap between the combinations of the spatial conditions during training and inference time, \ie, during training, all the spatial conditions extracted from the same image are fully aligned while the multiple spatial conditions accepted from users are not the case. 
User-provided conditions usually have multiple image sources, thus the extracted spatial conditions are not always aligned and sometimes have occlusion in the overlapping region, which requires the model to handle the spatial conditions in right arrangement according to the depth of the target scene.
To relieve the discrepancy, we collect a subset of unaligned data as shown in Figure~\ref{fig:unaligned_data}. 
To be specific, we utilize the images in Open Images dataset~\cite{openimages} and MSCOCO~\cite{coco} dataset which are rich in objects to make the synthetic data. 
Given an image and the mask of a foreground object, we recover the background image with the masked region using the inpainting tool~\cite{PowerPaint}. We discard images with too small or too large objects, and finally produce 0.44M images as supplementary unaligned training data. 

\vspace{0.1in}
\noindent{\textbf{Training Strategy.}} 
 When utilizing the unaligned data for training, we take the combination of spatial conditions for the foreground object and the inpainted background image together while treating the original image as target. During training, for the data with fully aligned spatial conditions, we randomly pick two conditions for each training sample; for the synthetic unaligned data, we randomly pick a condition for the foreground object and the background inpainted image respectively. We randomly drop all the conditions at a rate of 0.05 to enable classifier free guidance, and also randomly drop the textual prompts at a rate of 0.05 to let the model learn from pure spatial conditions only.

\section{Experiments}
We validate the effectiveness of AnyControl with Stable Diffusion~\cite{LDM} of version 1.5 on four types of conditions, including Edge~\cite{Canny}, Depth~\cite{MiDAS}, Segmentation~\cite{EntitySeg}, and Human Pose~\cite{Openpose}. We compare our AnyControl with state-of-the-art methods including Multi-ControlNet and Multi-Adapter, the versions of ControlNet~\cite{ControlNet} and T2I-Adapter~\cite{T2I-Adapter} which support multiple spatial conditions, as well as Uni-ControlNet~\cite{uni-controlnet}, Cocktail~\cite{Cocktail},  UniControl~\cite{unicontrol}, DiffBlender~\cite{DiffBlender} and CnC~\cite{CnC} with extensive qualitative and quantitative results. Implementation details including network structure, hyper-parameters of training and inference can be found in the supplementary material. 

\begin{figure}
    \centering
    \includegraphics[width=1.0\linewidth]{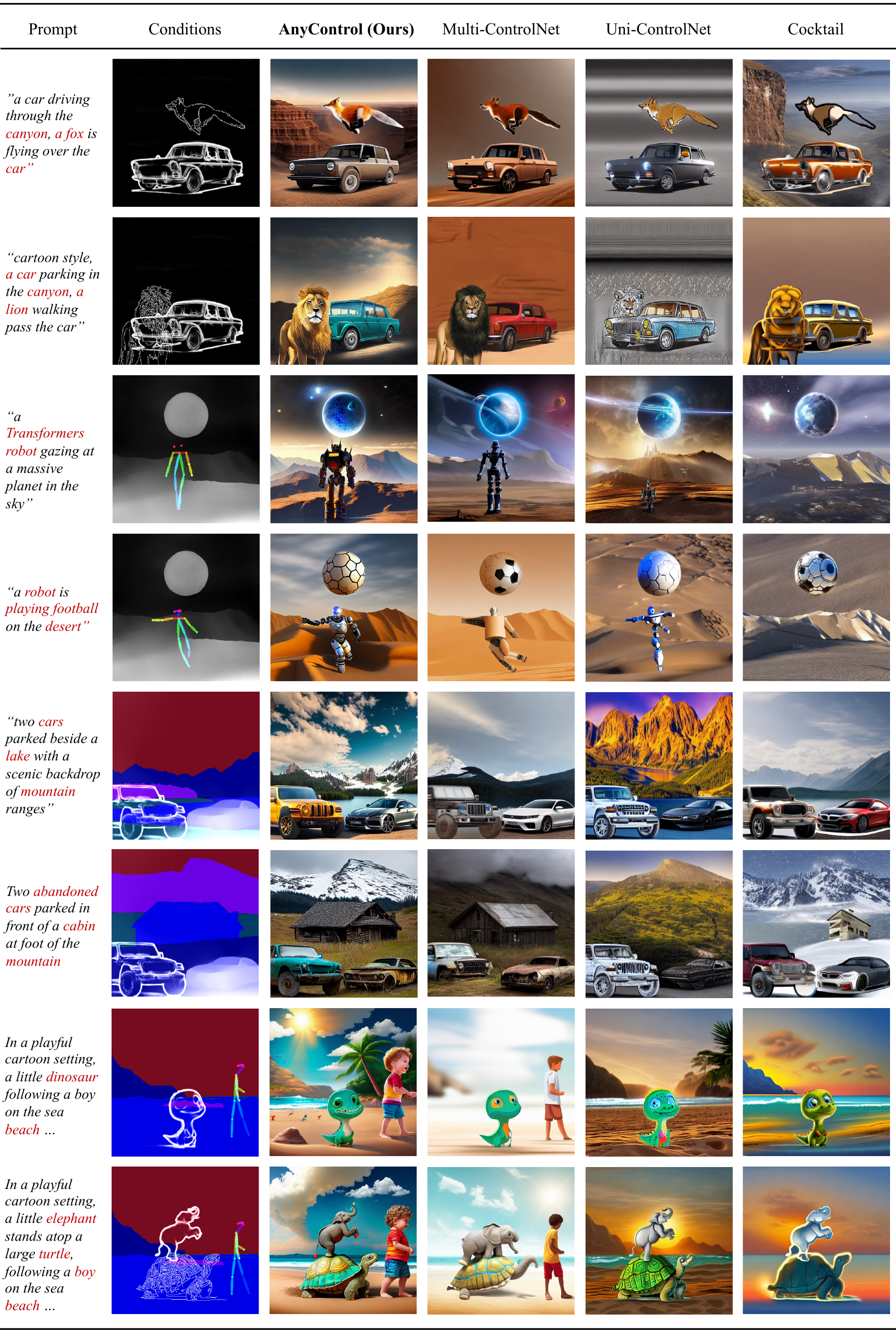}
    \caption{\textbf{Comparison on multi-control image synthesis.} 
    Multi-ControlNet adopts MoE design to process diverse conditions while Cocktail adopts the composition design by combining multiple conditions of the same type into one. 
    }
    \label{fig:main_results}
\end{figure}

\subsection{Qualitative Results}
In this section, we analyze input flexibility, spatial compatibility, and also the compatibility with text, style and color control. 

\begin{figure}[t]
    \centering
    \includegraphics[width=1.0\linewidth]{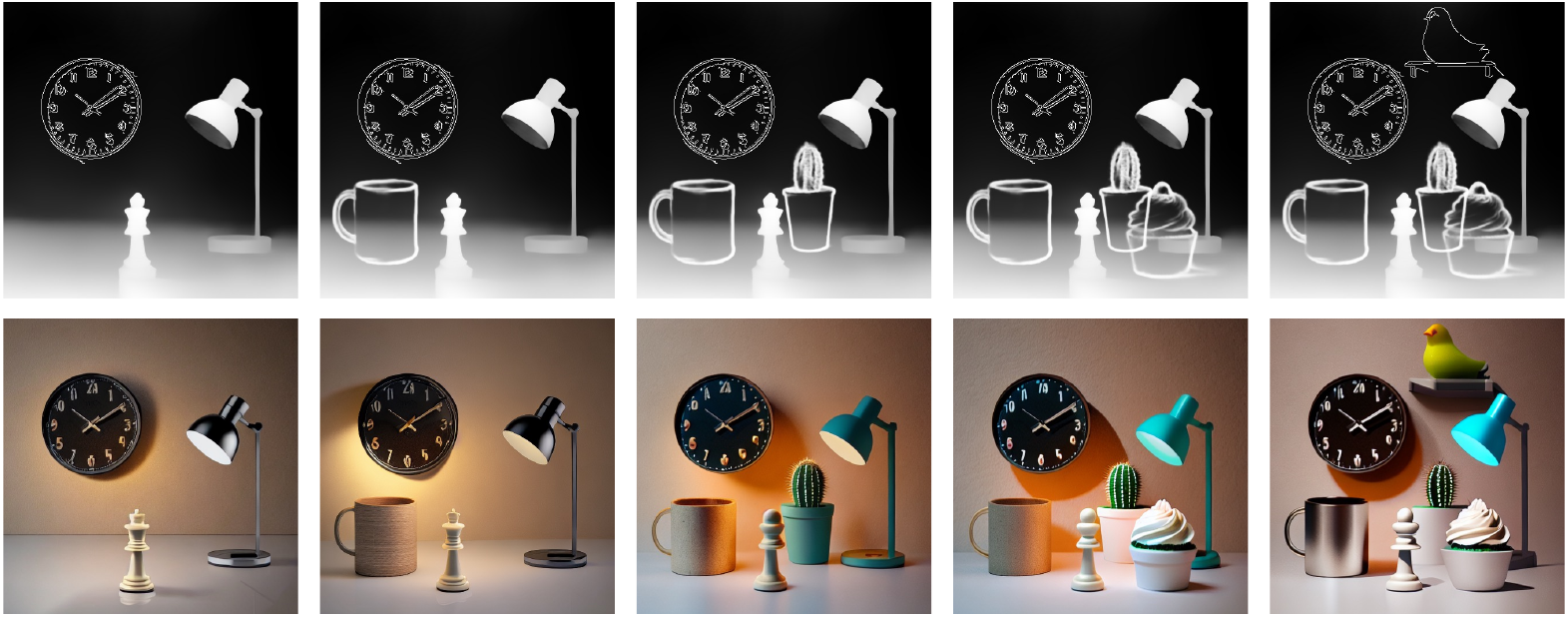}
    \vspace{-0.2in}
    \caption{\textbf{Varied-number and varied-type of input spatial conditions.}}
    \label{fig:many_conditions}
    \vspace{-0.2in}
\end{figure}

\vspace{0.1in}
\noindent{\textbf{Input Flexibility.}} 
There are three ways to process free combinations of spatial conditions from users: 1) MoE design (\ie, UniControl\cite{unicontrol}, Multi-ControlNet~\cite{ControlNet}); 2) Attention design (AnyControl); 3) Composition design, which merges spatial conditions of the same type into one image so that methods with fixed-length input channels can work smoothly. In MoE-based methods, the composition of different conditions is achieved by hand-crafted weighted summation. Instead, our AnyControl adopts attention mechanism to learn the composition weights dynamically, achieving superior performance on multi-control image synthesis as shown in Figure~\ref{fig:main_results} and Figure~\ref{fig:many_conditions}. In addition,``sticker'' artifact is observed in the results of Cocktail~\cite{Cocktail} with composition design.

\vspace{0.1in}
\begin{wrapfigure}{r}{0.5\textwidth}
    \centering
\vspace{-0.3in}
\includegraphics[width=0.48\textwidth]{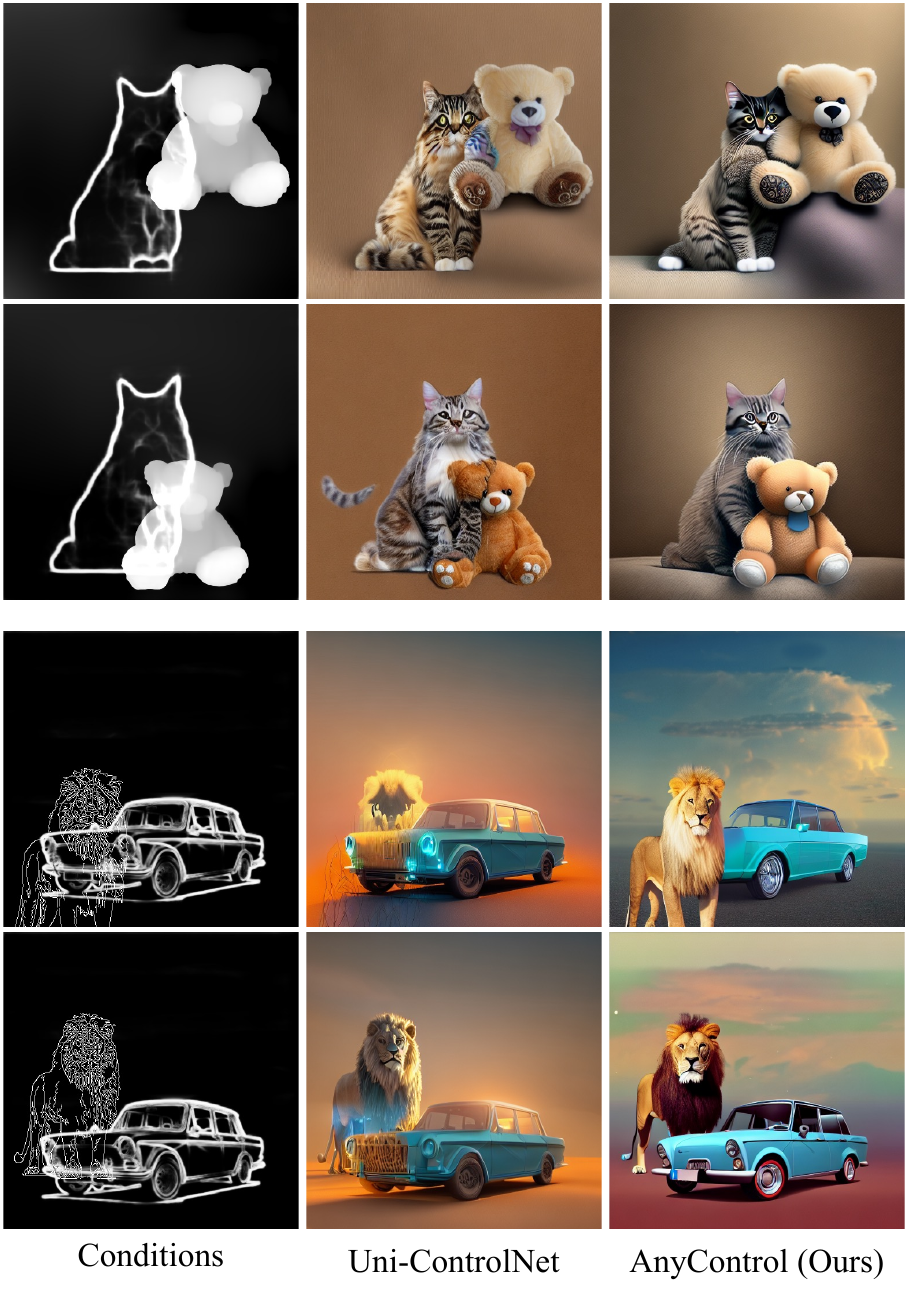}
\vspace{-0.1in}
\caption{\textbf{Spatial compatibility.} AnyControl is capable of inferring the relationships not only between conditions but also between the generated objects and the environment.}
\label{fig:occluded_cases} 
\vspace{-0.3in}
\end{wrapfigure}
\noindent{\textbf{Spatial Compatibility.}} 
In Figure~\ref{fig:main_results}, we 
provide
comparisons given various conditions with occlusion, demonstrating the superiority of AnyConrol in handling
complex multi-control synthesis. Blending issues are difficult to avoid in previous
methods~\cite{ControlNet, uni-controlnet, Cocktail} with trivial design on multi-control combination.  

Figure~\ref{fig:occluded_cases} further shows examples on AnyControl in dealing with relative spatial positions of two conditions, which generates high-quality results with
correct spatial relation between conditions.
For example, given different layout 
arrangements, the generated cat and teddy bear from AnyControl are always positioned in a natural way, rather than mixed within the occluded region. 
Another notable thing is that AnyControl shows a surprising ability of dealing with the interaction of generated objects and the corresponding environment. When placed at the same horizontal line, the cat and teddy bear are seated at the same plane, while the teddy bear sitting on a stage when its vertical position axis is raised up. These advantages contribute to the introduction of the Multi-Control Encoder, which strengthens the interactions between all control signals and consequently achieves a comprehensive understanding of complex user inputs.

\vspace{0.1in}
\noindent{\textbf{Text Compatibility.}} Textual prompts play an important role in controllable image synthesis as text is typically the primary communication means between human and T2I models, while spatial conditions work as auxiliary roles in providing the fine-grained information. Therefore, while guided by spatial conditions, maintaining the compatibility with textual prompts is essential. However, existing methods commonly prioritize the response to spatial conditions and miss important message in textual prompts. For instance, in the third row of Figure~\ref{fig:main_results}, other methods either totally neglect the ``Transformers Robot'' information so that fail to bind the concept with the input human pose, or only partially respond to ``Robot'' but drop ``Transformers'' information. On the contrary, our method responds to all important information in multi-modality user inputs and produces harmonious results.

\vspace{0.1in}
\noindent{\textbf{Compatibility with Style and Color Control.}} 
As a plug-and-play model, 
\begin{wrapfigure}{r}{0.6\textwidth}
\vspace{-0.25in}
\centering
\includegraphics[width=0.59\textwidth]{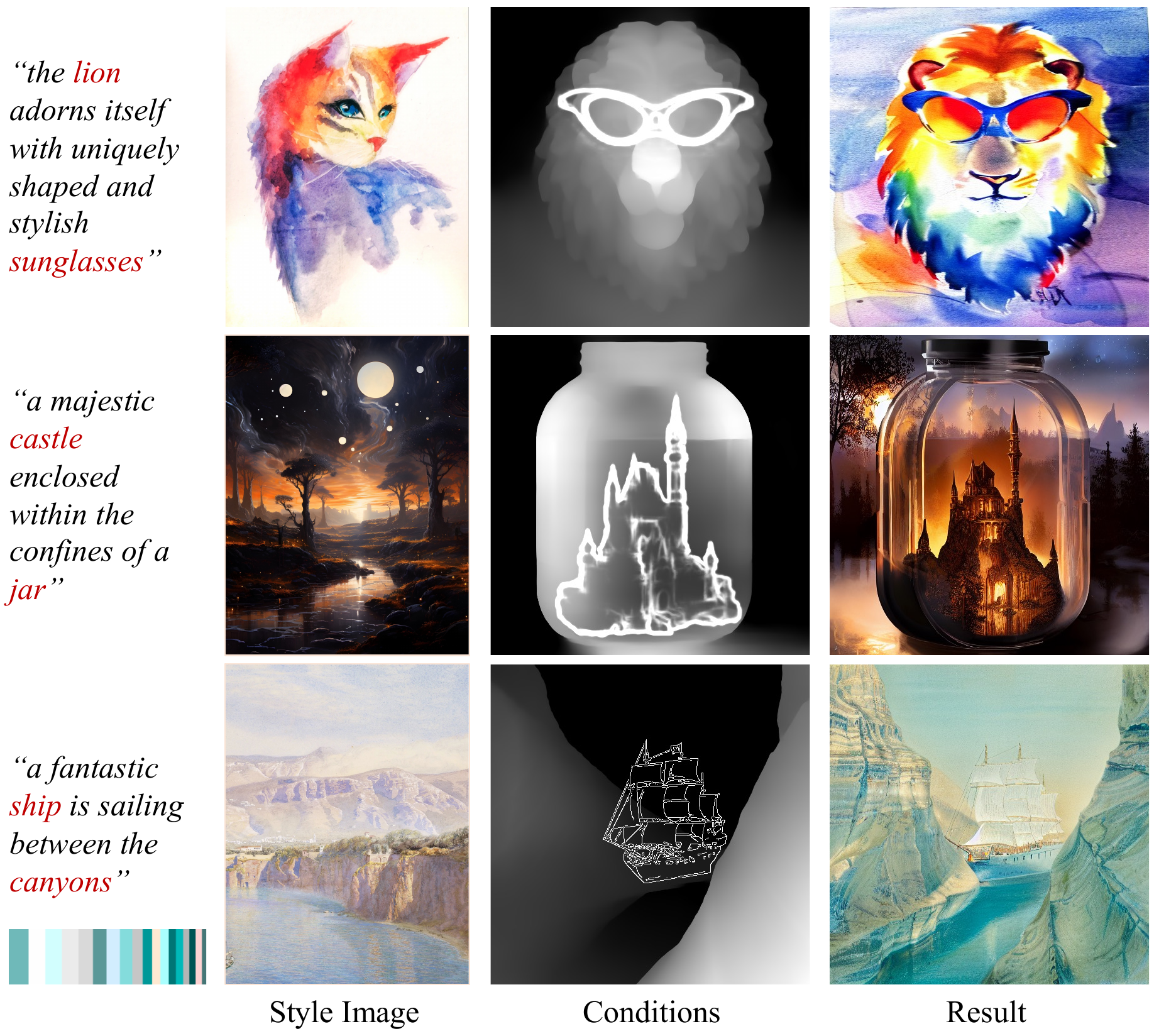}
\vspace{-0.1in}
\caption{\textbf{AnyControl with style and color controls.} The first two cases take style, depth and edge controls, while the last further takes color control.}
\vspace{-0.3in}
\label{fig:style_color} 
\end{wrapfigure}
AnyControl can be integrated with existing conditional generation methods in a convenient way. 
We take style and color controls as examples to demonstrate the effectiveness of AnyControl in the collaboration with other plug-and-play modules for wider applications. Specifically, we enhance AnyControl with decoupled cross-attention~\cite{IP-Adapter} in UNet to employ style and color controls. The compositional outcomes are visually depicted in Figure~\ref{fig:style_color}, revealing the generation of high-quality results that adhere to style, color and spatial constraints.

\subsection{Multi-Control Benchmark: \benchmark}
Most existing methods evaluate multi-control image synthesis on MSCOCO validation set with totally spatio-aligned conditions extracted from different algorithms. However, we argue that evaluation on the dataset with well-aligned multi-control conditions cannot reflect the ability of methods to handle occluded multiple conditions in practical applications, given that the user provided conditions are typically collected from diverse sources which are not aligned. 

Therefore, we construct an \textbf{U}naligned \textbf{M}ulti-control benchmark based on MSCOCO validation set, short for \textbf{COCO-UM}, for a more effective evaluation on multi-control image synthesis.  The construction pipeline is similar to that used in unaligned data synthesis described in Section~\ref{sec:training}. That is, we decompose an image into the background image and foreground image outlined by an object mask, and then recover the background image through inpainting tools~\cite{PowerPaint}. 

Additionally, after obtaining the recovered background image, we remove the bad cases with low image quality, such as cases with new generated object within the hole region rather than filled by background scene. Finally, we construct an occluded multi-control dataset with 1480 samples. 

\begin{table*}[t!]
 \begin{center}
 \caption{\textbf{Comparisons of multi-control image synthesis on \benchmark with existing multi-control methods.} Results with the best and the second best performance are highlighted in \topone{red} and \toptwo{blue}, respectively. 
  }
  \vspace{-0.2in}
	\label{tab:coco_um}
		\resizebox{\textwidth}{!}{
            \setlength{\tabcolsep}{2.7mm}
			\begin{tabular}{l|c|c|c|c|c|c}
				\toprule[1.5pt]
                    \textbf{Methods} & \textbf{FID}$\downarrow$ & \textbf{CLIP}$\uparrow$ & \makecell{\textbf{Depth (RMSE$\downarrow$)}} & 
                    \makecell{\textbf{Seg. (mPA$\uparrow$)}} &
                    \makecell{\textbf{Edge (RMSE$\downarrow$)}}  & \makecell{\textbf{Pose (mAP$\uparrow$)}}\\
			\midrule
                Multi-ControlNet~\cite{ControlNet} & 55.95 & 24.80 & \topone{17.81} & \toptwo{42.78} & 47.35 &  15.69 \\
                
                Multi-Adapter~\cite{T2I-Adapter} & 51.67 & \toptwo{25.73} & 24.08 &  32.62 & \toptwo{44.86} & 6.28 \\
   
                Uni-ControlNet~\cite{uni-controlnet} & 55.28 & 24.48 & 20.57 & 41.10 & \topone{43.28} & \toptwo{18.40} \\
                
                UniConrol~\cite{unicontrol} & 57.46 & 24.72 & 23.69 & 34.58 & 45.48 & 4.06 \\
                Cocktail~\cite{Cocktail} & \toptwo{47.39} & 25.33 & - & 31.74 & - & 12.16 \\
                
                % DiffBlender~\cite{DiffBlender} &  \\
                % CnC~\cite{CnC} & \\
                \midrule
                AnyControl (Ours) & \topone{44.28} & \topone{26.41} & \toptwo{18.00} & \topone{43.34} & 45.25 & \topone{18.81} \\
                
			\bottomrule[1.5pt]
			\end{tabular}}
	\end{center}
\end{table*}

\begin{table*}[t!]
\begin{center}\vspace{-0.2in}
\caption{\textbf{Comparisons of single control  with existing single- and multi-control methods on COCO-5K.} Results with the best and the second best performance are highlighted in \topone{red} and \toptwo{blue}, respectively. 
 }
 \vspace{-0.2in}
	\label{tab:single_control}
             \setlength{\tabcolsep}{2.0mm}
		\resizebox{\textwidth}{!}{
			\begin{tabular}{l|ccc|ccc|cc|cc}
				\toprule[1.5pt]
	            \multicolumn{1}{c|}{ \multirow{2}{*}{\textbf{Methods}}} &
				\multicolumn{3}{c|}{\textbf{Depth}} & 
				\multicolumn{3}{c|}{\textbf{Segmentation}} & 
				\multicolumn{2}{c|}{\textbf{Edge}} & 
				\multicolumn{2}{c}{\textbf{Pose}} \\
				& FID $\downarrow$ & CLIP $\uparrow$ & RMSE $\downarrow$ & FID $\downarrow$ & CLIP $\uparrow$ & mPA $\uparrow$ & FID $\downarrow$ & CLIP $\uparrow$ & FID $\downarrow$ & CLIP $\uparrow$ \\
			\midrule
			\rowcolor{lightgray} \multicolumn{11}{l}{\textbf{Single-Control Methods}} \\
			\midrule
                ControlNet~\cite{ControlNet} & \toptwo{19.80} & 25.30 & \topone{13.86} & 20.39 & 25.46 & 45.68 & \toptwo{16.16} & 25.34 & 26.15 & 25.39 \\
                T2I-Adapter~\cite{T2I-Adapter} & 20.08 & 25.67 & 15.62 & 20.95 & 24.91 & 33.67 & 18.76 & 25.25 & \toptwo{24.94} & \toptwo{25.76} \\
                \midrule
                \rowcolor{lightgray} \multicolumn{11}{l}{\textbf{Multi-Control Methods}} \\
			\midrule
                Uni-ControlNet~\cite{uni-controlnet} & 20.09 & 25.25 & 15.93 & 22.96 & 25.11 & 35.56 & 17.51 & 25.18  & 26.61 & 24.86 \\
                UniControl~\cite{unicontrol} & 20.67 & 25.51 & \toptwo{14.07} &  \toptwo{19.73} & \topone{26.11} & \toptwo{45.77} & 16.69 & 25.15 & 27.90 & 25.27 \\
                Cocktail~\cite{Cocktail} & - & - & - & 26.02 & 25.28 & 33.10 & \topone{15.70} & 24.91 & 28.36 & 25.27 \\
                DiffBlender~\cite{DiffBlender} & 21.32 & 25.59 & 18.07 & - & - & - & 21.48 & \toptwo{25.57} & 30.33 & 25.30 \\
                CnC~\cite{CnC} & 20.42 & \topone{26.95} & 17.57 & - & - & - & - & - & -& -\\
                \midrule
                AnyControl (Ours) & \topone{18.04} & \toptwo{25.98} & 15.48 & \topone{18.89} & \toptwo{26.02}& \topone{48.73}  & 18.89  & \topone{25.88} & \topone{24.12}
& \topone{25.99} \\
			\bottomrule[1.5pt]
			\end{tabular}}
	\end{center}
 \vspace{-0.3in}
\end{table*}

\vspace{-0.1in}
\subsection{Quantitative Evaluation}
For thorough quantitative evaluation, we employ various evaluation metrics, including FID~\cite{FID} for the restructured image quality, CLIP-Score~\cite{CLIP-Score} for the alignment with textual prompt. For the condition fidelity, we adopt RMSE for depth map and edge map, mPA and mAP for segmentation map and pose map, respectively.

\vspace{0.1in}
\noindent{\textbf{Multi-Control Synthesis Evaluation.}} 
We evaluate multi-control synthesis on \benchmark. As depicted in Table~\ref{tab:coco_um}, our method outperforms other multi-control methods on FID and CLIP-Score by a large margin. This remarkable achievement signifies that AnyControl is capable of processing complex combinations of multiple spatial conditions, and generates high-quality harmonious  results well aligned with the textual prompts and spatial conditions.

\vspace{0.1in}
\noindent{\textbf{Single-Control Synthesis Evaluation.}}  For comprehensive evaluation, we also conduct the comparisons on each single condition as tabulated in Table~\ref{tab:single_control}. To be specific, we evaluate the single control synthesis on the full validation set of
MSCOCO, that is, COCO-5K since there is no occlusion in single control scenario. 
Overall, AnyControl outperforms existing single- and multi-control methods on 
most metrics, illustrating the superiority of 
our method to existing methods.

\vspace{0.1in}
\noindent{\textbf{Ablation Study on Unaligned Data.}} 
Unaligned data is provided to solve the gap between the alignment of input conditions during training and inference. That is, in training, conditions are totally aligned; while, in testing, the control
\begin{wraptable}{r}{0.4\textwidth}
\vspace{-0.4in}
  \centering
  % \vspace{-0.35in}
  \caption{\textbf{Ablation study of training with and without unaligned data on \benchmark.}}
  \vspace{0.1in}
  \label{tab:unaligned_data}
  \setlength{\tabcolsep}{3.2mm}
  \scalebox{0.7}{
  \begin{tabular}{l|cc}
    \toprule[1.5pt]
    Metric & FID$\downarrow$ & CLIP$\uparrow$ \\
     \midrule
     w/o unaligned data & 52.10 & 25.62 \\
       w unaligned data & 44.28 & 26.40 \\
    \bottomrule[1.5pt]
  \end{tabular}
  \vspace{-0.4in}
  }
\vspace{-0.2in}
\end{wraptable}
signals from users contributing to a whole image are almost not aligned at all.
In Table~\ref{tab:unaligned_data}, we provide comparisons on FID and CLIP-Score of AnyControl
trained with and without unaligned data. As illustrated, a large improvement on FID and CLIP-Score is observed through the data expansion on occluded cases.
The introduction of unaligned data during training strengthens AnyControl in modeling complex multi-control synthesis, especially the occluded cases.

\subsection{Discussion}
Although the input number of spatial conditions is not limited in AnyControl, we 
\begin{wrapfigure}{r}{0.5\textwidth}
\vspace{-0.25in}
\centering
\includegraphics[width=0.49\textwidth]{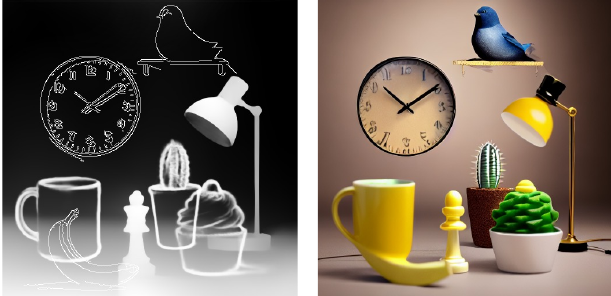}
\vspace{-0.1in}
\caption{\textbf{Miss-blending issue under too many spatial conditions.}}
\vspace{-0.25in}
\label{fig:discussion} 
\end{wrapfigure}
observe the miss-blending issue as shown in Figure~\ref{fig:discussion}, when the number of spatial
conditions is overlarge, such as 8 in this case. 
The possible reasons are as follows: 1) the limited ability of CLIP text encoder in understanding complex textual prompts with numerous concepts; 2) Too many visual tokens in cross-attention transformer block results in a decrease in the accuracy of softmax, and thus weaken AnyControl in precise multi-control understanding. We leave this issue as future work.

\section{Conclusion}

In conclusion, we propose a multi-control image synthesis framework based on the public T2I model to address the limitations of existing methods in accommodating diverse inputs, handling relationships among spatial conditions, and maintaining the compatibility with textual prompts. \method supports free combination of versatile control signals, which develops a Multi-Control Encoder that enables holistic understanding of multi-modal user inputs. We achieve this through employing alternating multi-control fusion and alignment blocks united by a set of query tokens. This approach enables AnyControl model complex relationships among diverse control signals and extract a multi-control embedding with compatible information. Our method produces high-quality natural outcomes, positioning it as a state-of-the-art solution for multi-condition image generation. The advancements introduced by \method contribute to the broader goal of enhancing controllable image synthesis and pushing the boundaries of T2I generation.

\vspace{0.3in}
\noindent{\textbf{Acknowledgement.}} This project is supported by the National Key R\&D Program of China (No. 2022ZD0161600)

\bibliographystyle{splncs04}
\bibliography{main}

\begin{thebibliography}{10}
\providecommand{\url}[1]{\texttt{#1}}
\providecommand{\urlprefix}{URL }
\providecommand{\doi}[1]{https://doi.org/#1}

\bibitem{SpaText}
Avrahami, O., Hayes, T., Gafni, O., Gupta, S., Taigman, Y., Parikh, D., Lischinski, D., Fried, O., Yin, X.: Spatext: Spatio-textual representation for controllable image generation. In: CVPR (2023)

\bibitem{MultiDiffusion}
Bar-Tal, O., Yariv, L., Lipman, Y., Dekel, T.: Multidiffusion: Fusing diffusion paths for controlled image generation  (2023)

\bibitem{MaskSketch}
Bashkirova, D., Lezama, J., Sohn, K., Saenko, K., Essa, I.: Masksketch: Unpaired structure-guided masked image generation. In: CVPR (2023)

\bibitem{Canny}
Canny, J.: A computational approach to edge detection. TPAMI  \textbf{PAMI-8}(6),  679--698 (1986)

\bibitem{Openpose}
Cao, Z., Simon, T., Wei, S.E., Sheikh, Y.: Realtime multi-person 2d pose estimation using part affinity fields. In: CVPR (2017)

\bibitem{LayoutDiffuse}
Cheng, J., Liang, X., Shi, X., He, T., Xiao, T., Li, M.: Layoutdiffuse: Adapting foundational diffusion models for layout-to-image generation. arXiv preprint arXiv:2302.08908  (2023)

\bibitem{ZestGuide}
Couairon, G., Careil, M., Cord, M., Lathuili{\`e}re, S., Verbeek, J.: Zero-shot spatial layout conditioning for text-to-image diffusion models. In: ICCV. pp. 2174--2183 (2023)

\bibitem{make-a-scene}
Gafni, O., Polyak, A., Ashual, O., Sheynin, S., Parikh, D., Taigman, Y.: Make-a-scene: Scene-based text-to-image generation with human priors. In: ECCV (2022)

\bibitem{TextualInversion}
Gal, R., Alaluf, Y., Atzmon, Y., Patashnik, O., Bermano, A.H., Chechik, G., Cohen-Or, D.: An image is worth one word: Personalizing text-to-image generation using textual inversion. arXiv preprint arXiv:2208.01618  (2022)

\bibitem{ResNet}
He, K., Zhang, X., Ren, S., Sun, J.: Deep residual learning for image recognition. In: CVPR (2016)

\bibitem{CLIP-Score}
Hessel, J., Holtzman, A., Forbes, M., Bras, R.L., Choi, Y.: Clipscore: A reference-free evaluation metric for image captioning. arXiv preprint arXiv:2104.08718  (2021)

\bibitem{FID}
Heusel, M., Ramsauer, H., Unterthiner, T., Nessler, B., Hochreiter, S.: Gans trained by a two time-scale update rule converge to a local nash equilibrium. NeurIPS  \textbf{30} (2017)

\bibitem{DDPM}
Ho, J., Jain, A., Abbeel, P.: Denoising diffusion probabilistic models. NeurIPS  \textbf{33},  6840--6851 (2020)

\bibitem{CFG}
Ho, J., Salimans, T.: Classifier-free diffusion guidance. arXiv preprint arXiv:2207.12598  (2022)

\bibitem{Cocktail}
Hu, M., Zheng, J., Liu, D., Zheng, C., Wang, C., Tao, D., Cham, T.J.: Cocktail: Mixing multi-modality control for text-conditional image generation. In: NeurIPS (2023)

\bibitem{Composer}
Huang, L., Chen, D., Liu, Y., Shen, Y., Zhao, D., Zhou, J.: Composer: Creative and controllable image synthesis with composable conditions. arXiv preprint arXiv:2302.09778  (2023)

\bibitem{SSMG}
Jia, C., Luo, M., Dang, Z., Dai, G., Chang, X., Wang, M., Wang, J.: {SSMG:} spatial-semantic map guided diffusion model for free-form layout-to-image generation. arXiv preprint arXiv:2308.10156  (2023)

\bibitem{DiffBlender}
Kim, S., Lee, J., Hong, K., Kim, D., Ahn, N.: Diffblender: Scalable and composable multimodal text-to-image diffusion models. arXiv preprint arXiv:2305.15194  (2023)

\bibitem{openimages}
Kuznetsova, A., Rom, H., Alldrin, N., Uijlings, J., Krasin, I., Pont-Tuset, J., Kamali, S., Popov, S., Malloci, M., Kolesnikov, A., et~al.: The open images dataset v4: Unified image classification, object detection, and visual relationship detection at scale. IJCV  \textbf{128}(7),  1956--1981 (2020)

\bibitem{CnC}
Lee, J., Cho, H., Yoo, Y., Kim, S.B., Jeong, Y.: Compose and conquer: Diffusion-based 3d depth aware composable image synthesis. In: ICLR (2024)

\bibitem{BLIP2}
Li, J., Li, D., Savarese, S., Hoi, S.: Blip-2: Bootstrapping language-image pre-training with frozen image encoders and large language models. arXiv preprint arXiv:2301.12597  (2023)

\bibitem{GLIGEN}
Li, Y., Liu, H., Wu, Q., Mu, F., Yang, J., Gao, J., Li, C., Lee, Y.J.: Gligen: Open-set grounded text-to-image generation. In: CVPR (2023)

\bibitem{coco}
Lin, T.Y., Maire, M., Belongie, S., Hays, J., Perona, P., Ramanan, D., Doll{\'a}r, P., Zitnick, C.L.: Microsoft coco: Common objects in context. In: ECCV. pp. 740--755 (2014)

\bibitem{FreeControl}
Mo, S., Mu, F., Lin, K.H., Liu, Y., Guan, B., Li, Y., Zhou, B.: Freecontrol: Training-free spatial control of any text-to-image diffusion model with any condition. arXiv preprint arXiv:2312.07536  (2023)

\bibitem{T2I-Adapter}
Mou, C., Wang, X., Xie, L., Wu, Y., Zhang, J., Qi, Z., Shan, Y., Qie, X.: T2i-adapter: Learning adapters to dig out more controllable ability for text-to-image diffusion models. arXiv preprint arXiv:2302.08453  (2023)

\bibitem{GLIDE}
Nichol, A., Dhariwal, P., Ramesh, A., Shyam, P., Mishkin, P., McGrew, B., Sutskever, I., Chen, M.: Glide: Towards photorealistic image generation and editing with text-guided diffusion models. arXiv preprint arXiv:2112.10741  (2021)

\bibitem{EntitySeg}
Qi, L., Kuen, J., Shen, T., Gu, J., Guo, W., Jia, J., Lin, Z., Yang, M.H.: High quality entity segmentation. In: ICCV (2023)

\bibitem{unicontrol}
Qin, C., Zhang, S., Yu, N., Feng, Y., Yang, X., Zhou, Y., Wang, H., Niebles, J.C., Xiong, C., Savarese, S., et~al.: Unicontrol: A unified diffusion model for controllable visual generation in the wild. arXiv preprint arXiv:2305.11147  (2023)

\bibitem{Layoutllm-t2i}
Qu, L., Wu, S., Fei, H., Nie, L., Chua, T.S.: Layoutllm-t2i: Eliciting layout guidance from llm for text-to-image generation. In: MM (2023)

\bibitem{CLIP}
Radford, A., Kim, J.W., Hallacy, C., Ramesh, A., Goh, G., Agarwal, S., Sastry, G., Askell, A., Mishkin, P., Clark, J., et~al.: Learning transferable visual models from natural language supervision. In: ICML. pp. 8748--8763 (2021)

\bibitem{DALLE2}
Ramesh, A., Dhariwal, P., Nichol, A., Chu, C., Chen, M.: Hierarchical text-conditional image generation with clip latents. arXiv preprint arXiv:2204.06125  \textbf{1}(2), ~3 (2022)

\bibitem{MiDAS}
Ranftl, R., Lasinger, K., Hafner, D., Schindler, K., Koltun, V.: Towards robust monocular depth estimation: Mixing datasets for zero-shot cross-dataset transfer. TPAMI  \textbf{44}(3) (2022)

\bibitem{LDM}
Rombach, R., Blattmann, A., Lorenz, D., Esser, P., Ommer, B.: High-resolution image synthesis with latent diffusion models. In: CVPR (2022)

\bibitem{UNet}
Ronneberger, O., Fischer, P., Brox, T.: U-net: Convolutional networks for biomedical image segmentation. In: MICCAI. pp. 234--241 (2015)

\bibitem{DreamBooth}
Ruiz, N., Li, Y., Jampani, V., Pritch, Y., Rubinstein, M., Aberman, K.: Dreambooth: Fine tuning text-to-image diffusion models for subject-driven generation. In: CVPR (2023)

\bibitem{Imagen}
Saharia, C., Chan, W., Saxena, S., Li, L., Whang, J., Denton, E.L., Ghasemipour, K., Gontijo~Lopes, R., Karagol~Ayan, B., Salimans, T., et~al.: Photorealistic text-to-image diffusion models with deep language understanding. NeurIPS  \textbf{35},  36479--36494 (2022)

\bibitem{laion}
Schuhmann, C., Beaumont, R., Vencu, R., Gordon, C., Wightman, R., Cherti, M., Coombes, T., Katta, A., Mullis, C., Wortsman, M., et~al.: Laion-5b: An open large-scale dataset for training next generation image-text models. NeurIPS  \textbf{35},  25278--25294 (2022)

\bibitem{DDIM}
Song, J., Meng, C., Ermon, S.: Denoising diffusion implicit models. arXiv preprint arXiv:2010.02502  (2020)

\bibitem{InstanceDiffusion}
Wang, X., Darrell, T., Rambhatla, S.S., Girdhar, R., Misra, I.: Instancediffusion: Instance-level control for image generation (2024)

\bibitem{BoxDiff}
Xie, J., Li, Y., Huang, Y., Liu, H., Zhang, W., Zheng, Y., Shou, M.Z.: Boxdiff: Text-to-image synthesis with training-free box-constrained diffusion. In: ICCV (2023)

\bibitem{Law-diffusion}
Yang, B., Luo, Y., Chen, Z., Wang, G., Liang, X., Lin, L.: Law-diffusion: Complex scene generation by diffusion with layouts. In: ICCV (2023)

\bibitem{ReCO}
Yang, Z., Wang, J., Gan, Z., Li, L., Lin, K., Wu, C., Duan, N., Liu, Z., Liu, C., Zeng, M., et~al.: Reco: Region-controlled text-to-image generation. In: CVPR (2023)

\bibitem{IP-Adapter}
Ye, H., Zhang, J., Liu, S., Han, X., Yang, W.: Ip-adapter: Text compatible image prompt adapter for text-to-image diffusion models. arXiv preprint arXiv:2308.06721  (2023)

\bibitem{ControlNet}
Zhang, L., Rao, A., Agrawala, M.: Adding conditional control to text-to-image diffusion models. In: ICCV (2023)

\bibitem{uni-controlnet}
Zhao, S., Chen, D., Chen, Y.C., Bao, J., Hao, S., Yuan, L., Wong, K.Y.K.: Uni-controlnet: All-in-one control to text-to-image diffusion models. NeurIPS  \textbf{36} (2024)

\bibitem{LayoutDiffusion}
Zheng, G., Zhou, X., Li, X., Qi, Z., Shan, Y., Li, X.: Layoutdiffusion: Controllable diffusion model for layout-to-image generation. In: CVPR (2023)

\bibitem{PowerPaint}
Zhuang, J., Zeng, Y., Liu, W., Yuan, C., Chen, K.: A task is worth one word: Learning with task prompts for high-quality versatile image inpainting. arXiv preprint arXiv:2312.03594  (2023)

\end{thebibliography}

\clearpage
\newpage
\appendix
\noindent\textbf{\Large Appendix}
\bigskip

\begin{figure}
    \centering
    \includegraphics[width=1.0\linewidth]{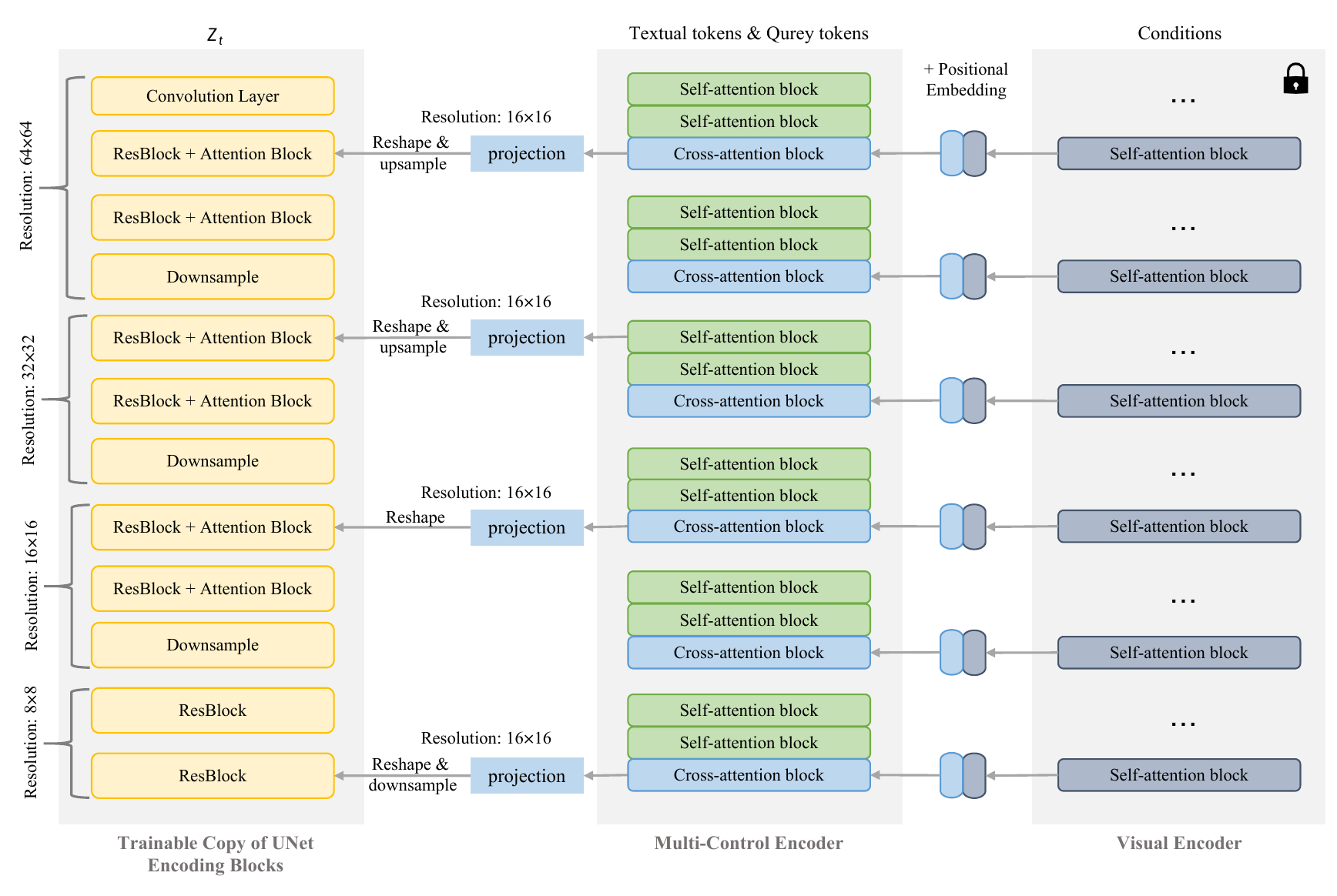}
    \vspace{-0.2in}
    \caption{Details of AnyControl and Multi-Control Encoder.}
    \label{fig:supp_network}
    \vspace{-0.2in}
\end{figure}

\section{Implementation Details}

\begin{figure}[t]
    \centering
    \includegraphics[width=1.0\linewidth]{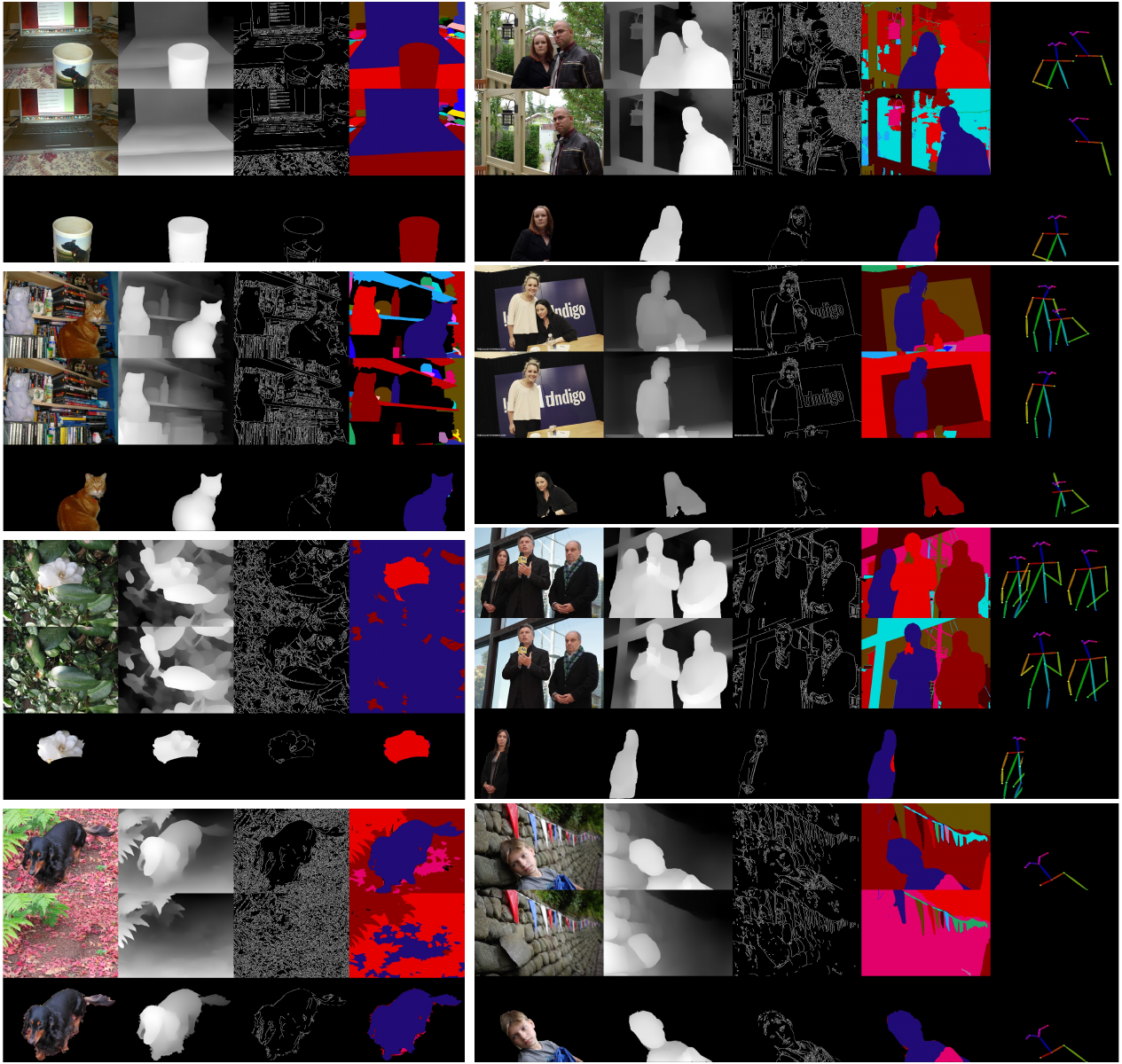}
    \vspace{-0.2in}
    \caption{More visualizations of unaligned data.}
    \label{fig:supp_unaligned_data}
    \vspace{-0.2in}
\end{figure}
\noindent{\textbf{Network.}} The detailed structure of our AnyControl is depicted in Figure~\ref{fig:supp_network}. We base Stable Diffusion of version 1.5 to build our AnyControl. Similar to ControlNet~\cite{ControlNet}, we make a trainable copy of the UNet encoding blocks for adapting to controlling information while freezing the pre-trained weights of Stable Diffusion model totally.  
In our Multi-Control Encoder, the number of query tokens is set to 256 enabling detailed controllable information extraction. The additional position embedding, with the same length as the query tokens, are shared by all input spatial conditions. We take the pre-trained weights of Q-Former~\cite{BLIP2} as the initialization for Multi-Control Encoder except for the query tokens and the additional position embedding, which are randomly initialized.  

\vspace{0.1in}
\noindent{\textbf{Hyper Parameters.}} We train AnyControl on 8 A100 GPU cards with a batch size of 8 on each GPU. We train the model for totally 90K iterations with a initial learning rate of 1e-5. During inference, we set the classifier-free guidance scale to 7.5. In all the experiments, we adopt DDIM~\cite{DDIM} sampler with 50 timesteps for all the compared methods.

\vspace{-0.1in}
\section{Unaligned Data}
\vspace{-0.1in}
During producing the synthetic unaligned dataset, we utilize the groudtruth object masks with the area ratio in $[0.1, 0.4]$ to outline the foreground object, while oversmall or overlarge objects will lead to undesired recovered background image. PowerPaint~\cite{PowerPaint} is a multi-task inpainting model supporting text-guided object inpainting, context-aware image inpainting as well as object removal. Here, we adopt the ``object removal'' mode for the unaligned data construction. More visualizations for synthetic unaligned data are in Figure~\ref{fig:supp_unaligned_data}.

\section{Hand-crafted Weight Adjustment}
\begin{figure}
    \centering
    \includegraphics[width=1.0\linewidth]{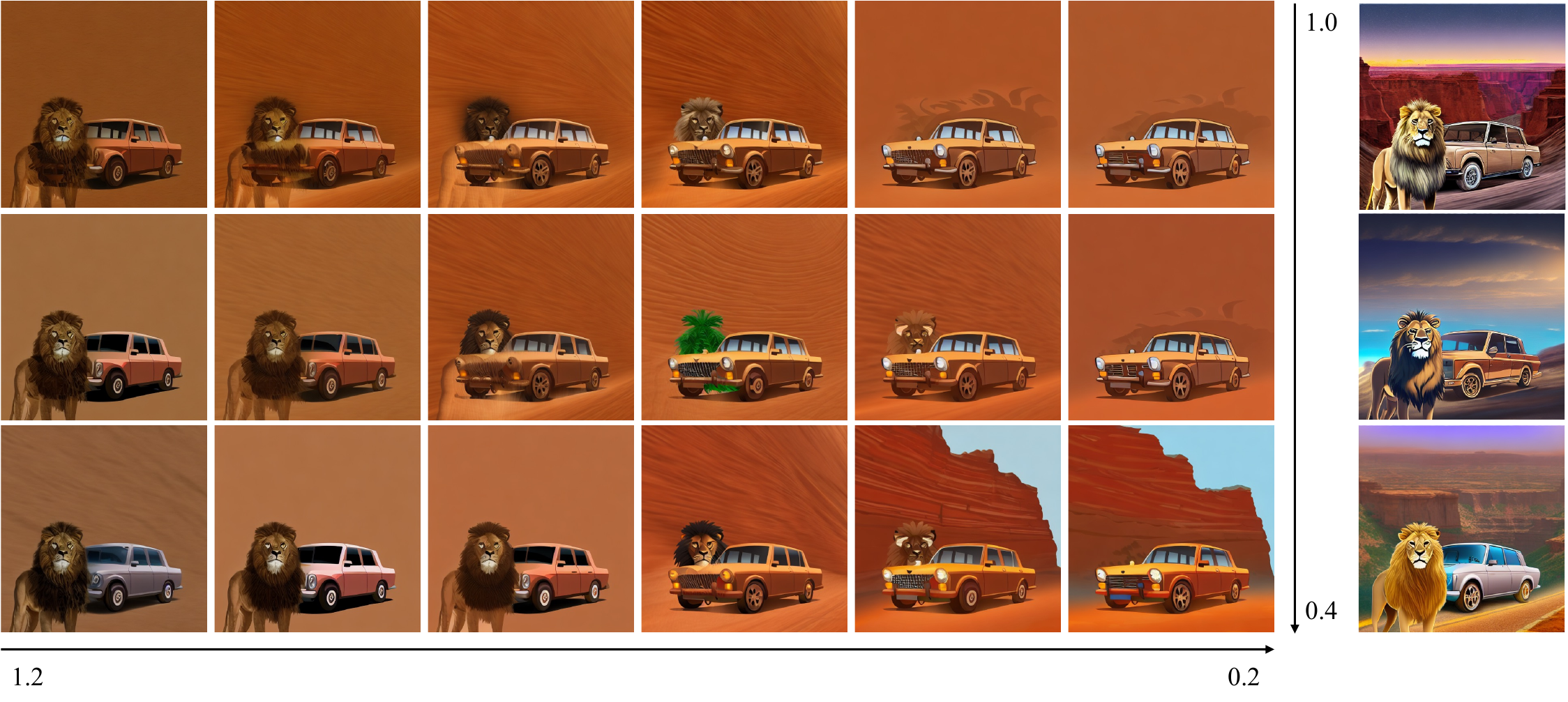}
    \vspace{-0.2in}
    \caption{Given prompt ``cartoon style, a car parking in the canyon, a lion walking pass the car'' and the edge conditions for a car as well as a lion, \textbf{left} shows hand-crafted weights adjustment for Multi-ControlNet~\cite{ControlNet}. X-axis and Y-axis represents the weight for lion and car conditions respectively. \textbf{Right} shows the results of AnyControl from three random seeds.}
    \label{fig:supp_weights}
    \vspace{-0.3in}
\end{figure}

As shown in Figure~\ref{fig:supp_weights}, multi-control methods with hand-crafted weights, \ie, Multi-ControlNet~\cite{ControlNet}, usually require a series of laborious weight adjustments according to the synthesized results while ours can automatically infer the combination weights and extracts unified multi-control embedding, thus producing harmonious results.

\section{Multi-level Visual Tokens}
Although the visual tokens from the last transformer block of the pre-trained visual encoder have 
already aggregated rich information, they are not sufficient 
\begin{wraptable}{r}{0.45\textwidth}
\vspace{-0.35in}
  \centering
  \caption{\textbf{Multi-level visual tokens.} We gradually enable the visual tokens from the deepest level to the shallowest level.
  }
  \label{tab:level_of_connections}
  \setlength{\tabcolsep}{1.2mm}
  \scalebox{0.7}{
  \begin{tabular}{l|cccccc}
    \toprule[1.5pt]
    Levels & 1 & 2 & 3 & 4 & 5 & 6 \\
     \midrule
     FID $\downarrow$ & 45.64 & 43.73 & 43.69 & 43.67 & 43.74 & 44.28 \\
     CLIP $\uparrow$ & 26.35 & 26.40 & 26.39 & 26.39 & 26.38 & 26.40 \\
    \bottomrule[1.5pt]
  \end{tabular}
  }
\end{wraptable}
to convey fine-grained 
controllable information. 
We conduct ablation experiments on the levels of used visual tokens from the visual encoder to the multi-control encoder. Table~\ref{tab:level_of_connections} demonstrate that integrating more visual tokens from middle layers increase FID and encounter performance saturation at 4-th level.

\section{More Qualitative Results}
\begin{figure}[t]
    \centering
    \includegraphics[width=1.0\linewidth]{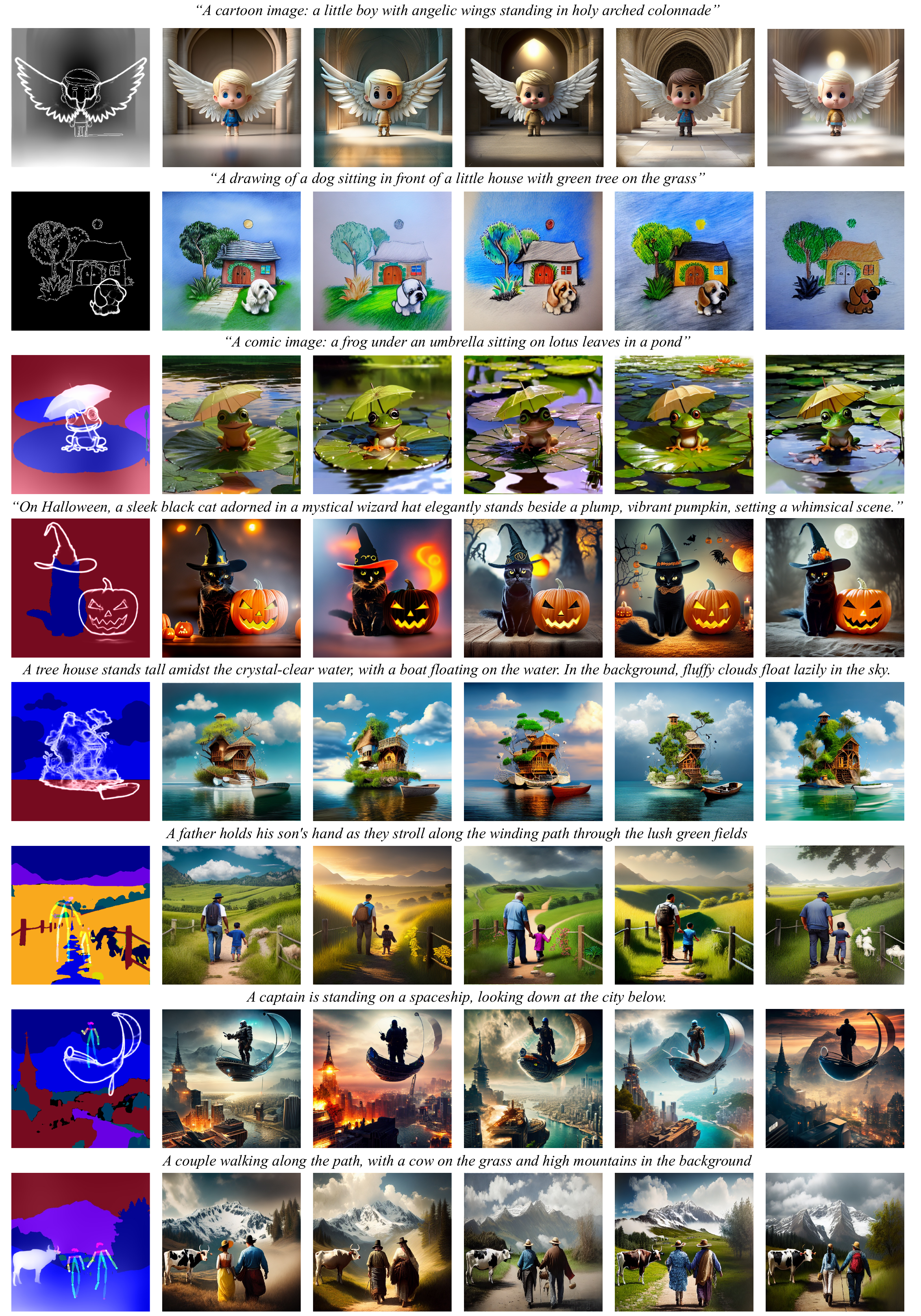}
    \caption{More visual results from AnyControl on multi-control image synthesis.}
    \label{fig:supp_multi_control}
\end{figure}

\begin{figure}[t]
    \centering
    \includegraphics[width=1.0\linewidth]{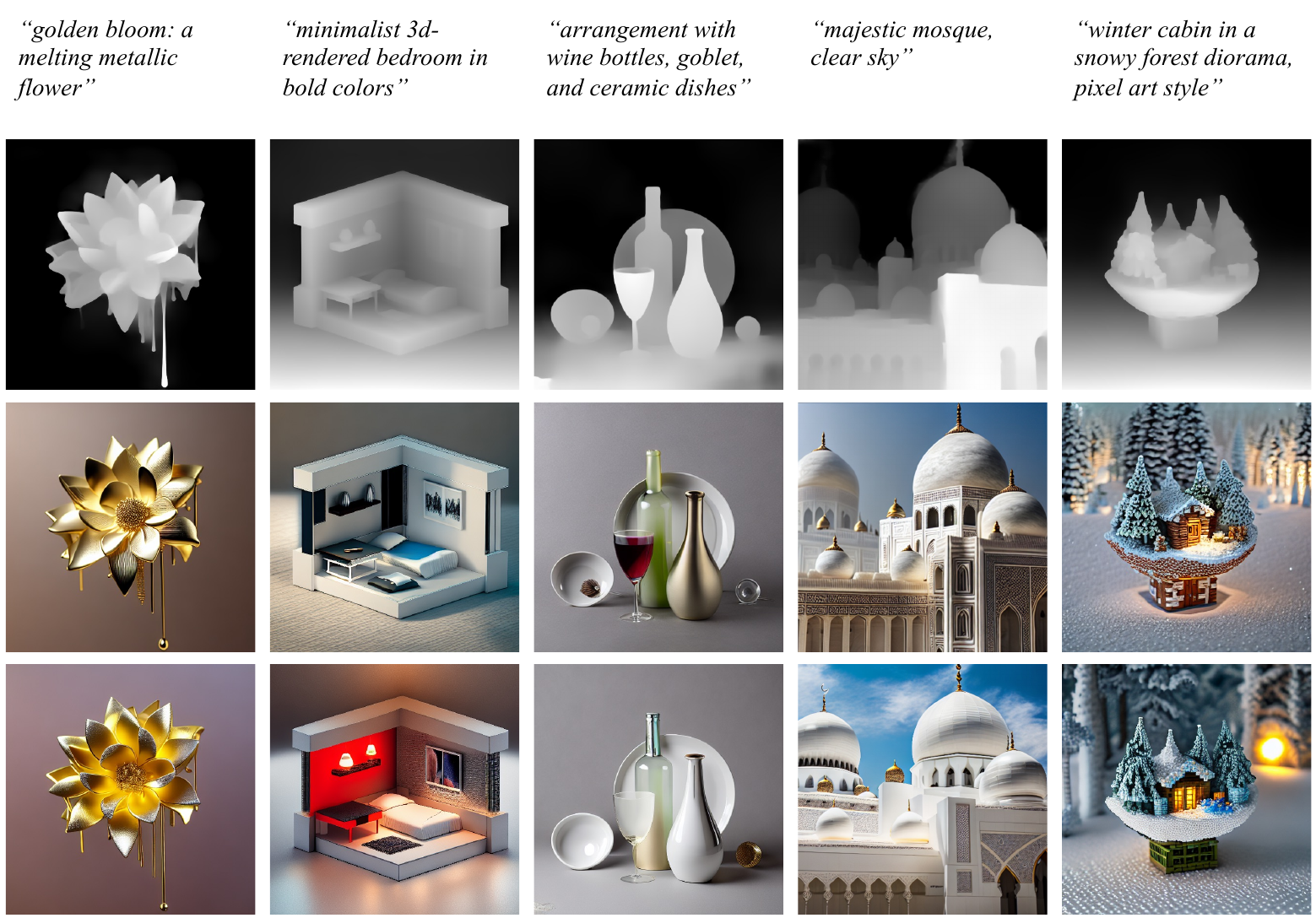}
    \caption{Visual results on depth controlled image synthesis.}
    \label{fig:supp_single_depth}
\end{figure}

\begin{figure}
    \centering
    \includegraphics[width=1.0\linewidth]{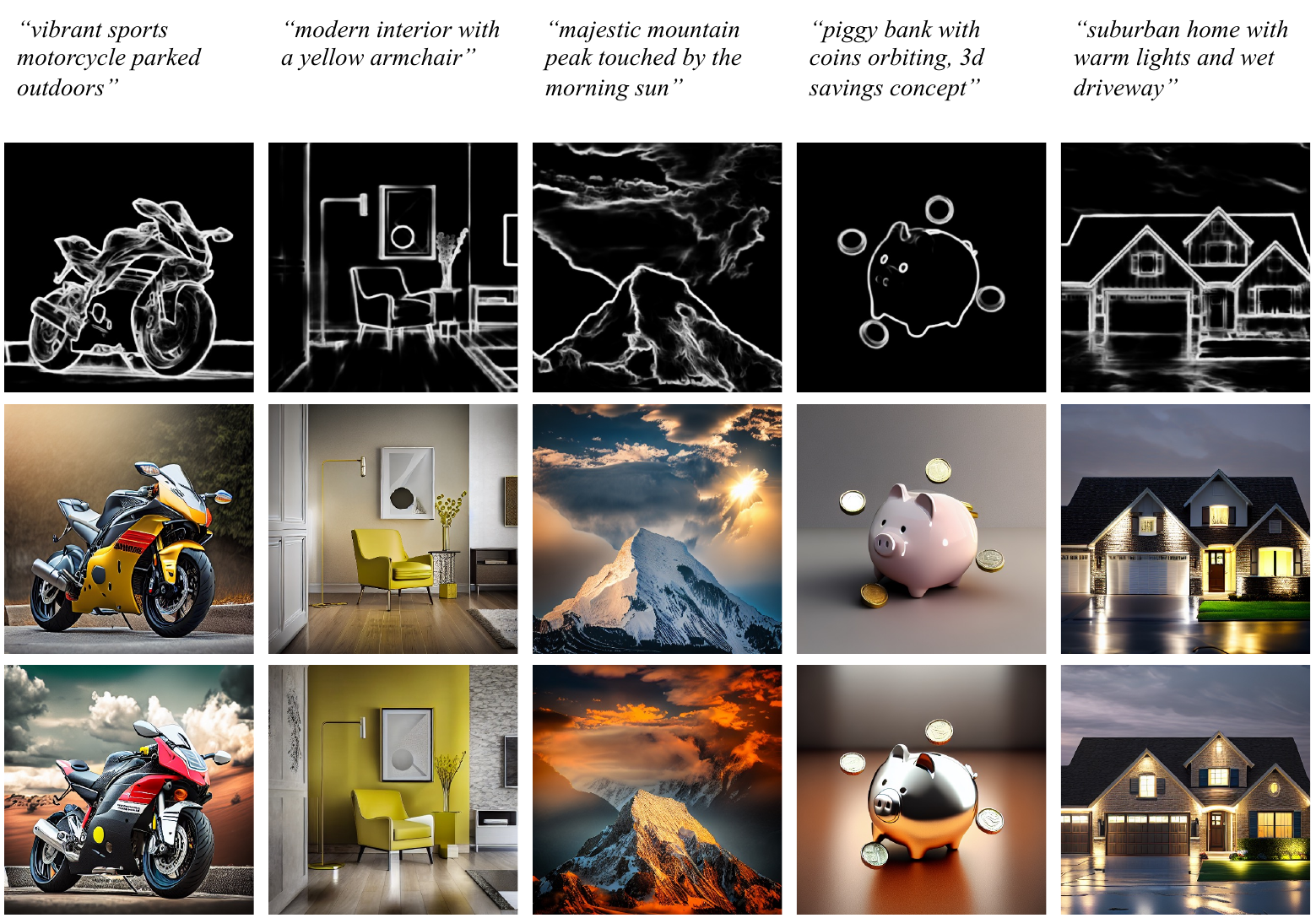}
    \caption{Visual results on edge controlled image synthesis.}
    \label{fig:supp_single_edge}
\end{figure}

\begin{figure}[t]
    \centering
    \includegraphics[width=1.0\linewidth]{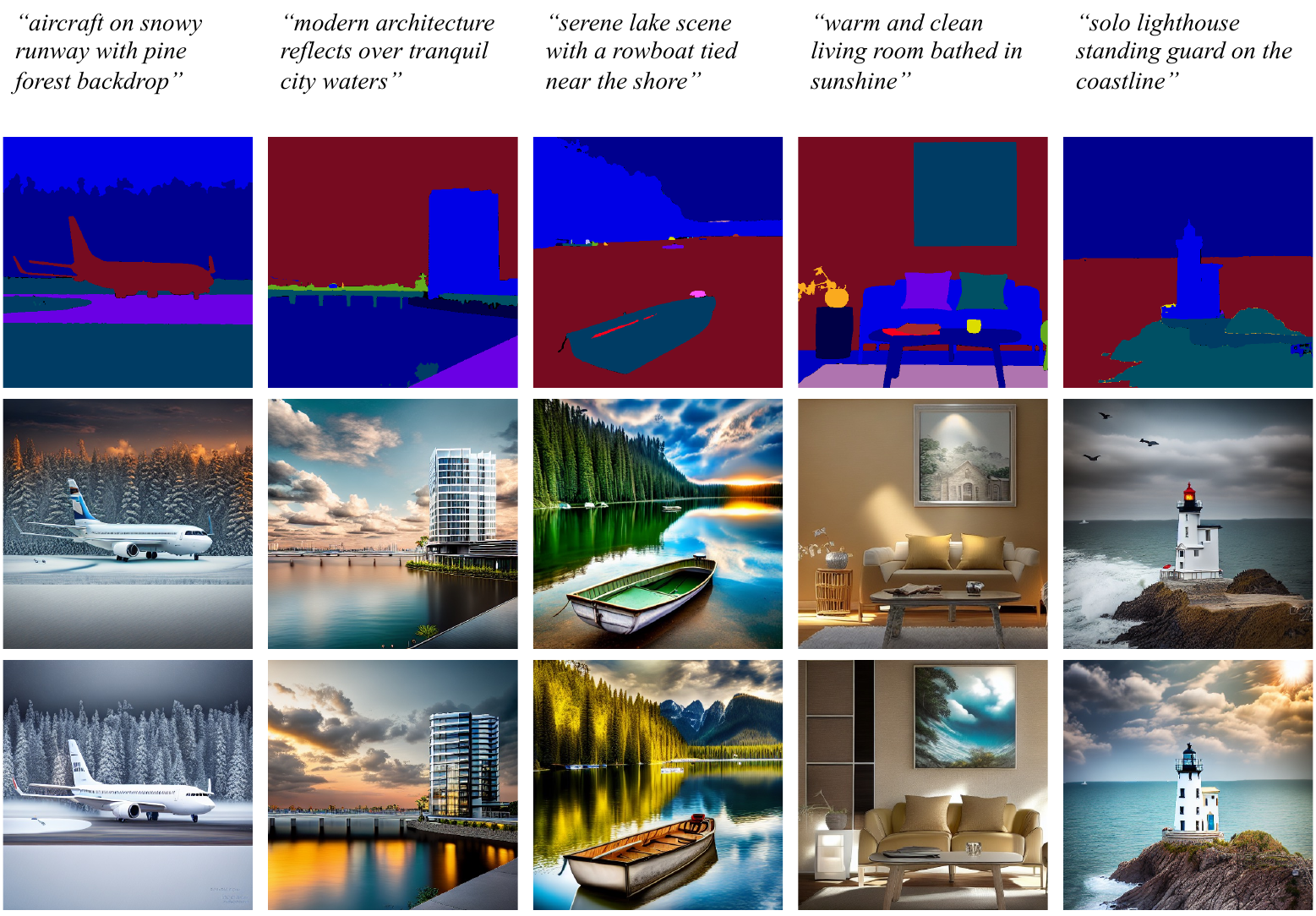}
    \caption{Visual results on segmentation controlled image synthesis.}
    \label{fig:supp_single_seg}
\end{figure}

\begin{figure}
    \centering
    \includegraphics[width=1.0\linewidth]{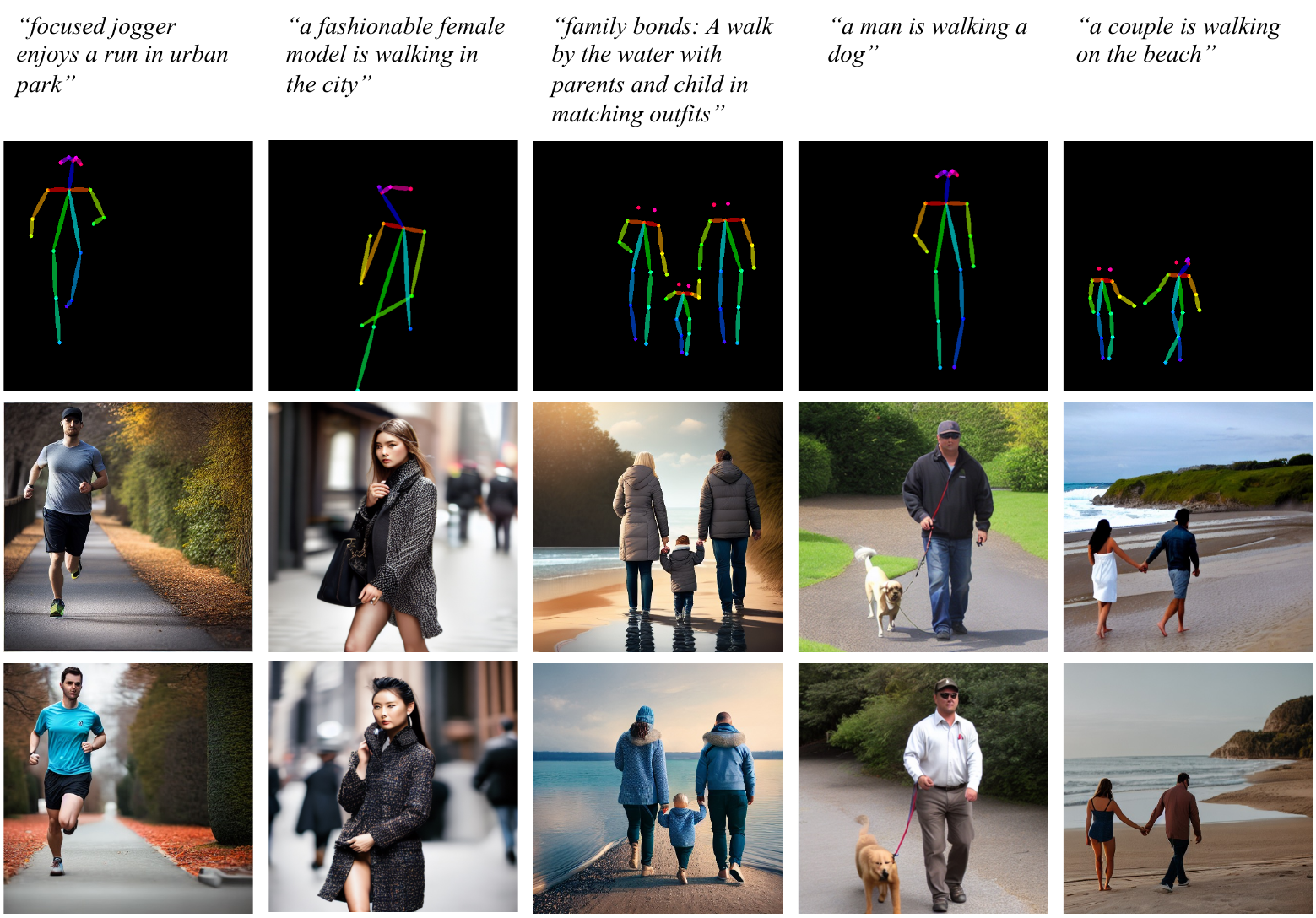}
    \caption{Visual results on human pose controlled image synthesis.}
    \label{fig:supp_single_pose}
\end{figure}

More qualitative results on multi-control synthesis are shown in Figure~\ref{fig:supp_multi_control}. Results of single-control synthesis including depth map, edge map, segmentation map and human pose are shown in Figure~\ref{fig:supp_single_depth} to Figure~\ref{fig:supp_single_pose} respectively.

\newpage

\newpage

\newpage

\end{document}